\documentclass[runningheads]{llncs}
\usepackage{graphicx}
\usepackage{amsmath,amssymb} 
\usepackage{color}
\usepackage[width=122mm,left=12mm,paperwidth=146mm,height=193mm,top=12mm,paperheight=217mm]{geometry}

\usepackage{bm}
\newcommand{\norm}[1]{\left\lVert#1\right\rVert} 
\newcommand{\bx}{\mathbf{x}}
\newcommand{\bb}{\mathbf{b}}
\DeclareMathOperator*{\argmin}{arg\,min}

\usepackage{array}
\newcolumntype{V}{>{\centering\arraybackslash} m{.43\linewidth} } 
\newcolumntype{W}{>{\centering\arraybackslash} m{.55 \linewidth}} 

\usepackage{color}

\usepackage{tabularx}
\usepackage{enumitem}

\usepackage[export]{adjustbox}

\usepackage{algcompatible,algorithm}

\usepackage{caption}

\begin{document}
\pagestyle{headings}
\mainmatter

\title{Robust Face Alignment \\Using a Mixture of Invariant Experts} 

\titlerunning{Robust Face Alignment Using a Mixture of Invariant Experts}

\authorrunning{O. Tuzel, T.K. Marks, and S. Tambe}

\author{Oncel Tuzel\inst{1} \and Tim K. Marks\inst{1} \and Salil Tambe\inst{2}}
\institute{Mitsubishi Electric Research Labs (MERL)\\
\email{onceltuzel@gmail.com, tmarks@merl.com}
\and Intel Corporation\\
\email{salil.tambe@intel.com}}

\maketitle


\begin{abstract}

Face alignment, which is the task of finding the locations of a set of facial landmark points in an image of a face, is useful in widespread application areas. Face alignment is particularly challenging when there are large variations in pose (in-plane and out-of-plane rotations) and facial expression. To address this issue, we propose a cascade in which each stage consists of a mixture of regression experts. Each expert learns a customized regression model that is specialized to a different subset of the joint space of pose and expressions. The system is invariant to a predefined class of transformations (e.g., affine), because the input is transformed to match each expert's prototype shape before the regression is applied. We also present a method to include deformation constraints within the discriminative alignment framework, which makes our algorithm more robust. Our algorithm significantly outperforms previous methods on publicly available face alignment datasets.
%
\end{abstract}

\section{Introduction} \label{sec:intro}

Face alignment refers to finding the pixel locations of a set of predefined facial landmark points (e.g., eye and mouth corners) in an input face image. It is important for many applications such as human-machine interaction, videoconferencing, gaming, and animation, as well as numerous computer vision tasks including face recognition, face tracking, pose estimation, and expression synthesis. Face alignment is difficult due to large variations in factors such as pose, expression, illumination, and occlusion. 

\subsection{Previous work}


Great strides have been made in the field of face alignment since the Active Shape Model (ASM)~\cite{cootes1995active} and Active Appearance Model (AAM)~\cite{cootes2001active} were first proposed. AAM-based face alignment methods proposed since then include~\cite{sauer2011accurate,sung2009adaptive,tzimiropolous2013aam}. 
To handle wider variations in pose, multi-view AAM and ASM models \cite{romdhani1999multiview,cootes2002viewAAM,Asthana2011ICCV3DposeNorm} explicitly model and predict the head pose, e.g., by learning a different deformable model for each of several specific pose ranges~\cite{cootes2002viewAAM,Asthana2011ICCV3DposeNorm}. Another line of research involves multi-camera AAMs, in which an AAM is simultaneously fitted to images of a face captured by multiple cameras~\cite{cootes2000coupledAAM,hu2004multicamAAM}.
Like ASMs and AAMs, Constrained Local Models (CLMs)~\cite{cristinacce2006feature,cristinacce2007boosted,zhou2013exemplar,smith2012joint} have explicit joint constraints on the landmark point locations (e.g., a subspace shape model) that constrain the positions of the landmarks with respect to each other. Building on CLMs, \cite{tzimiropoulos2014gauss} propose the Gauss-Newton Deformable Part Model \mbox{(GN-DPM)}, which uses Gauss-Newton optimization to jointly fit an appearance model and a global shape model. 

Recently, much of the focus in face alignment research has shifted toward discriminative methods~\cite{tuzel2008learning,xiong2013supervised,asthana2014incremental,liu2009discriminative,kazemi2011face,saragih2011deformable,dollar2010cascaded}. These methods learn an explicit regression that directly maps the features extracted at the facial landmark locations to the face shape (e.g., the locations of the landmarks) \cite{xiong2013supervised,asthana2014incremental,ren2014face,kazemi2014one,cao2014face,tzimiropoulos2015POCR}. In Project-Out Cascaded Regression (PO-CR)~\cite{tzimiropoulos2015POCR}, the regression is performed in a subspace orthogonal to facial appearance variation. To cope with inaccurate initialization,~\cite{yan2013multiHyp} begin a regression cascade at multiple initial locations and combine the results. Tree-based regression methods \cite{ren2014face,kazemi2014one,cao2014face,cootes2012robust,tuzel2008learning} are also gathering interest due to their speed. In \cite{ren2014face}, a set of local binary features are learned using a random forest regression  to jointly learn a linear regression function for the final estimation, while \cite{kazemi2014one} utilize a gradient boosting tree algorithm to learn an ensemble of regression trees. Software libraries such as~\cite{menpo14} implement a wide range of face alignment methods.

In the Supervised Descent Method (SDM)~\cite{xiong2013supervised}, a cascade of regression functions operate on extracted SIFT features to iteratively estimate facial landmark locations. An extension of SDM, called Global SDM (GSDM)~\cite{xiong2015globalSDM}, partitions the parameter space into regions of similar gradient direction, and uses the result from the previous frame of video to determine which region's model to use in the current frame.  Unlike our method, which takes individual test images as input, GSDM is a tracking method that requires a video sequence. Other methods that report results only on video input include~\cite{Xiao_2015_ICCV_Workshops}.

A variety of recent face alignment methods incorporate deep neural networks, including deep regression networks~\cite{Zhang2015iccv} and coarse-to-fine neural network approaches~\cite{Zhou2013iccvWorkshop,Zhang2014eccvCFAN}. A different coarse-to-fine approach is taken by~\cite{Zhu2015cvprCFSS}. Other recent variations on face alignment research include methods that are specially designed to handle partially occluded faces~\cite{Burgos-Artizzu_2013_cofw,yu2014consensus}.

\subsection{Our approach}
 
 \begin{figure}[!tb]
  \centering
  \begin{tabular}{V|p{4pt}W}
\includegraphics[width=.35\columnwidth,trim=.5in .6in .5in .5in, clip]{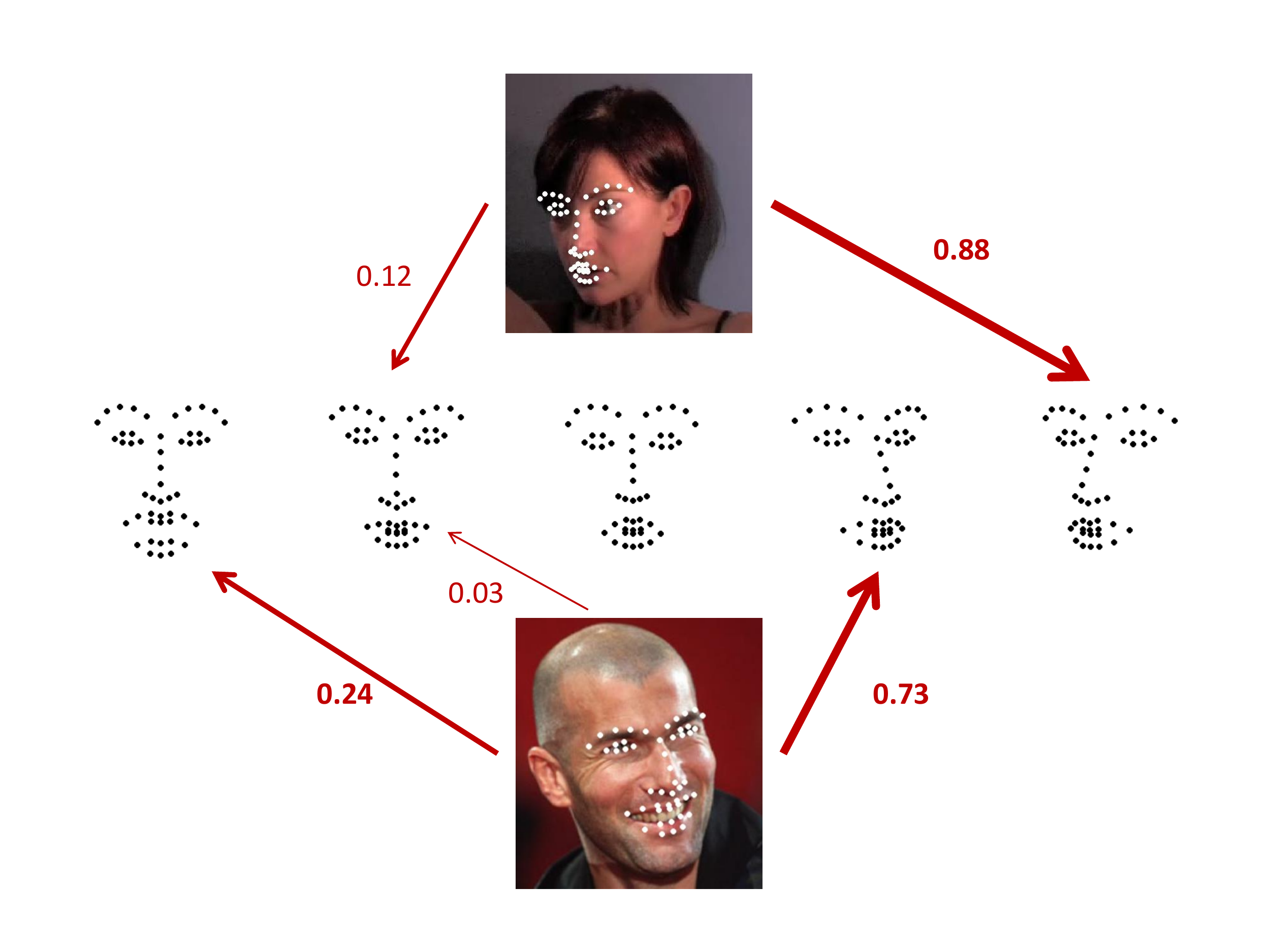}
&&\includegraphics[width=.5\columnwidth]{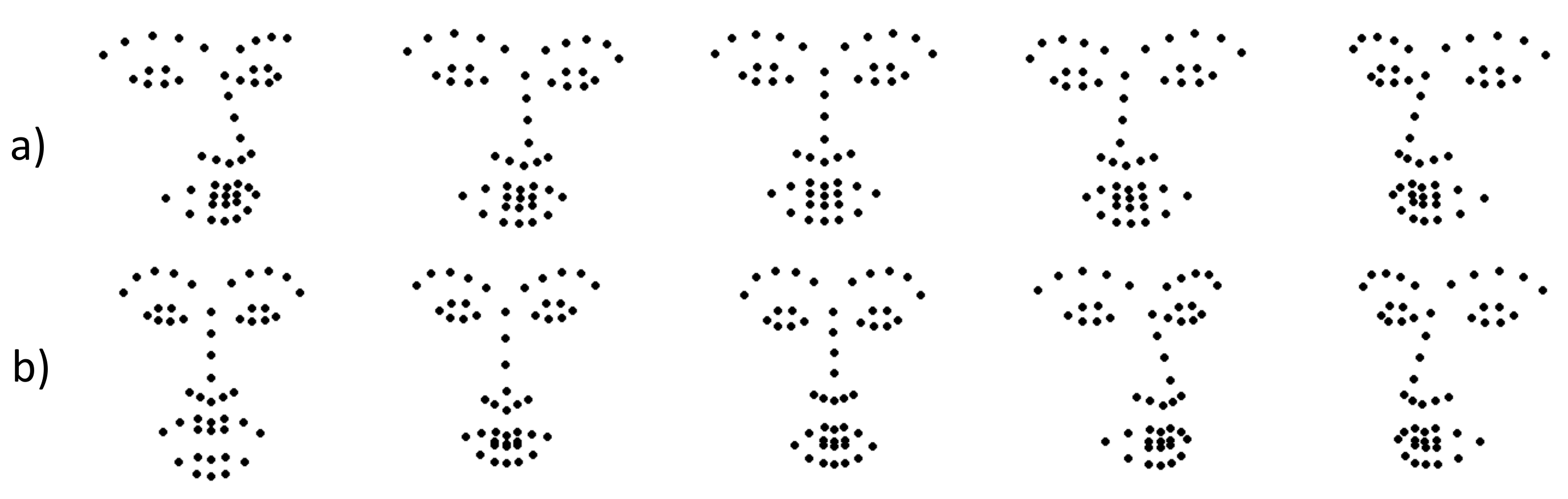}\\[-4pt]
  \end{tabular}
  \caption{{\it Left:} Each expert specializes in a subset of the possible poses and facial expressions. Arrows show the assignment weights of each image's landmark point configuration to the 5 experts. {\it Right:} Cluster evaluation. a)~Euclidean cluster centers. b)~Affine-invariant cluster centers. Affine-invariant clustering accounts for both the pose variations and the facial expressions.} 
\label{fig:MixtureAndClusters}
\end{figure}

Our method is related to SDM~\cite{xiong2013supervised} in that we also perform a cascade of regressions on SIFT features that are computed at the currently estimated landmark locations. However, our method improves upon SDM in a number of ways. 
In SDM, the same linear regression function must work across all possible variations in facial expressions and pose, including both in-plane and out-of-plane head rotations. In addition to requiring a large and varied training dataset, this forces the learned regression functions to be too generic, thereby limiting accuracy. We address this shortcoming in two ways. First, we propose a transformation invariance step at each level of the cascade, prior to the regression step, which makes our method invariant to an entire class of transformations. (Here, we choose the class of 2D affine transformations.) As a result, our regression functions do not need to correct for such global changes in pose and face shape, enabling them to be fine tuned to handle the remaining, smaller variations in landmark locations.

To further improve robustness to variations in pose and expression, at each stage of the cascade we replace the linear regression from SDM by a mixture of experts~\cite{rao1997MOEregression}. In our cascade, each stage is a mixture of experts, where each expert is a regression specialized to handle a subset of the possible face shapes (e.g., a particular region of the joint space of face poses and expressions). As illustrated in Fig.~\ref{fig:MixtureAndClusters} (left), each expert corresponds to a different prototype face shape. This improves alignment significantly, especially when the training dataset is biased towards a certain pose (e.g., frontal). 

Unlike alignment methods based on parametric shape models (such as AAM, ASM, and CLM), SDM has no explicit global constraints to jointly limit the locations of multiple landmark points. Our method addresses this limitation simply, by penalizing deviations of landmark locations from each expert's prototype face shape. We accomplish this in the regression framework by extending the feature vector to include the difference between the prototype landmark locations and the currently estimated landmark locations, weighted by a scalar that determines the rigidity of the model. This global regularization of the face shape prevents feature points from drifting apart.
\\[-16pt]



\paragraph{Contributions}
In summary, we propose a robust method for real-time face alignment which we call {\bf M}ixture of {\bf I}nvariant E{\bf x}perts (MIX). Novel elements include:
\begin{itemize}[nosep]
\item A transformation invariance step, before each stage of regression, which makes our method invariant to a specified class of transformations. (In this study, we choose the class of 2D affine transformations.)
\item A simple extension to the feature vectors that enables our regressions to penalize deviations of feature locations from a prototypical face shape.
\item A mixture-of-experts regression at each stage of the cascade, in which each expert regression function is specialized to align a different subset of the input data (e.g., a particular range of expressions and poses).
\item A novel affine-invariant clustering algorithm to learn the prototype shapes used in the mixture model.
\end{itemize}

These novel elements enable our method to achieve precise face alignment on a wide variety of images. 
We perform exhaustive tests on the 300W~\cite{sagonas2013CVPRlabels,sagonas2013ICCV300w} and AFW~\cite{zhu2012face} datasets, comparing with eight recent methods:
Coarse-to-Fine Auto-encoder Networks (CFAN)~\cite{Zhang2014eccvCFAN}, ensemble of regression trees (TREES)~\cite{kazemi2014one}, Coarse-to-Fine Shape Searching (CFSS)~\cite{Zhu2015cvprCFSS}, SDM~\cite{xiong2013supervised}, its incrementally learned adaptation CHEHRA~\cite{asthana2014incremental}, GN-DPM~\cite{tzimiropoulos2014gauss}, \mbox{Fast-SIC} (an AAM method trained on ``in-the-wild'' images) \cite{tzimiropolous2013aam}, and  \mbox{PO-CR} \cite{tzimiropoulos2015POCR}. We demonstrate that the proposed method significantly outperforms these previous state-of-the-art approaches.


\section{Supervised Descent Method} \label{sec:SDM}

We now describe the Supervised Descent Method (SDM)~\cite{xiong2013supervised}, which is related to our method, while introducing notation that we will use throughout the paper. Let $I$ be an input face image, and let $\mathbf{x}$ be the $2p \times 1$ vector of $p$ facial landmark locations in image coordinates. At each of the $p$ landmark locations in $\bx$, we extract a $d$-dimensional feature vector. In this paper, we use SIFT features~\cite{lowe2004distinctive} with $d=128$. Let $\bm{\phi}(I,\mathbf{x})$ be the $pd \times 1$ consolidated feature vector, which is a concatenation of the $p$ feature descriptors extracted from image $I$ at the landmark locations $\bx$. 

Given a current estimate, $\mathbf{x}_k$, of the landmark locations in image $I$, SDM formulates the alignment problem as finding an update vector $\Delta \mathbf{x}$ such that the features computed at the new landmark locations $\mathbf{x}_k + \Delta \mathbf{x}$ better match the features computed at the ground-truth landmark locations $\hat{\mathbf{x}}$ in the face image. The corresponding error can be written as a function of the update vector $\Delta \bx$:
\begin{equation}
	f(\mathbf{x}_k + \Delta \bx) = \bigl \| \bm{\phi}(I,\mathbf{x}_k + \Delta \mathbf{x})- \hat{\bm{\phi}} \bigr \|^2,
\label{eq:SDM}
\end{equation}
where we define $\hat{\bm{\phi}} = \bm{\phi}(I,\hat{\bx})$. This function $f$ could be minimized by Newton's method. 
%
%
%
The Newton step is given by 
\begin{equation}
\Delta \bx = -\mathbf{H}^{-1} \mathbf{J}_f = -2\mathbf{H}^{-1} \mathbf{J}_{\bm{\phi}}  \bigl [ \bm{\phi}_k - \hat{\bm{\phi}} \bigr ], 
\label{eq:newton}
\end{equation}
where $\mathbf{H}$ is the Hessian matrix of $f$, $\mathbf{J}_f$ and $\mathbf{J}_{\bm{\phi}}$ represent the Jacobian with respect to $\bx$ of $f$ and ${\bm{\phi}}$, respectively, and we define $\bm{\phi}_k = \bm{\phi}(I,\bx_k).$ The Hessian and Jacobian in~\eqref{eq:newton} are evaluated at $\bx_k$,
but we have omitted the argument $\bx_k$ to emphasize the dependence on $\bm{\phi}_k.$ 
In SDM, \eqref{eq:newton}~is approximated by the multivariate linear regression 
\begin{equation}
\Delta \bx = \mathbf{W}_k \bm{\phi}_k + \mathbf{b}_k,
\label{eq:regression}
\end{equation}
in which coefficients $\mathbf{W}_k$ and bias $\mathbf{b}_k$ do not depend on $\bx_k$. 

In SDM~\cite{xiong2013supervised}, a cascade of $K$ linear regressions $\{\mathbf{W}_k, \mathbf{b}_k\}$, where $k = 1, \ldots, K$, are learned using training data. Face alignment is achieved by sequentially applying the learned regressions to features computed at the landmark locations output by the previous stage of the cascade:
\begin{equation}
\bx_{k+1} = \bx_{k} + \mathbf{W}_{k}\bm{\phi}_{k} + \mathbf{b}_{k}.
\end{equation}
To learn the regressions $\{\mathbf{W}_k, \mathbf{b}_k\},$ the $N$ face images in the training data are augmented by repeating every training image $M$ times, each time perturbing the ground-truth landmark locations by a different random displacement. For each image $I_i$ in this augmented training set \mbox{$(i = 1, \ldots, M\!N),$} with ground-truth landmark locations $\hat{\bx}_i$, we displace the landmarks by random displacement $\Delta \hat{\bx}_i$.
The first regression function ($k=1$) is learned by minimizing the L2-loss function
\begin{equation}
\{\mathbf{W}_k, \mathbf{b}_k\} = 
	\arg \min_{\mathbf{W}, \mathbf{b}} \sum_{i=1}^{M\!N} \| \Delta \hat{\bx}_i - \mathbf{W} \bm{\phi}(I_i,\hat{\bx}_i - \Delta \hat{\bx}_i) - \mathbf{b}\|^2 .
\end{equation}
For training the later regressions $\{\mathbf{W}_k, \mathbf{b}_k\}_{k=2, \ldots, K},$ rather than using a random perturbation, the target $\Delta \hat{\bx}_i$ is the residual after the previous stages of the regression cascade.

\section{Mixture of invariant experts} \label{sec:MoE}

In this section, we present our model. Our model significantly improves upon the alignment accuracy and robustness of SDM by introducing three new procedures: a transformation invariance step before each stage of regression, learned deformation constraints on the regressions, and the use of a mixture of expert regressions rather than a single linear regression at each stage of the cascade.

\subsection{Transformation invariance}\label{sec:transformationInvariance}

In order for the regression functions in SDM~\cite{xiong2013supervised} to learn to align facial landmarks for any face pose and expression, the training data must contain sufficiently many examples of faces covering the entire space of possible variations. Although being able to align faces at any pose is a desired property, learning such a function requires collecting (or synthesizing) training data containing all possible face poses. In addition, the learning is a more difficult task when there are large variations in the training set, and hence either a sufficiently complex regression model (functional form and number of features) is required, or the alignment method will compromise accuracy in order to align all these poses. As a general rule, increased model complexity leads to poorer generalization performance. This suggests that a simpler or more regularized model, which learns to align faces for a limited range of poses, would perform better for those poses than would a general alignment model that has been trained on all poses.  As a simple example, consider a regression function that is trained using a single upright face image versus one trained using multiple in-plane rotations of that face image. In the former case, the regression function must have a root for the upright pose, whereas in the latter case, the regression function must have a root for every in-plane rotation. 

Our goal with transformation invariance is to train each regression on a smaller set of poses, while still being able to align faces in an arbitrary pose. To do so, we apply a transformation invariance step prior to each stage's regression function. We first construct a prototype shape, $\bar{\bx}$, which contains the mean location of each landmark point across all of the training data (after uniform scaling and translation transformations have been applied to each training image to make them all share a canonical location and scale).

In this paper, we choose affine transformations as our class of transformations for invariance, although one could also use our method with a different class of  transformations. At each stage $k$ of regression, we find the affine transformation $\mathbf A_k$ that transforms the landmark locations $\bx_k$ that were estimated by the previous stage of regression so as to minimize their sum of squared distances to the prototype landmark locations, $\bar{\bx}$:
\begin{equation}
	\mathbf{A}_{k} = \argmin_{\mathbf{A} \in \mathcal A} \norm{\mathbf{A}(\bx_{k}) - \bar{\bx}}^2,
	\label{eq:affine}
\end{equation}
where $\mathcal{A}$ denotes the set of all affine transformations.  Next, we use the transformation $\mathbf{A}_{k}$ to warp the input image $I$ and the landmark locations into the prototype coordinate frame: 
$I' = \mathbf{A}_{k} (I)$, and $\bx'_{k} = \mathbf{A}_{k} (\bx_{k})$. Note that  we slightly abuse notation here by using the same affine transformation operator $\mathbf{A}_{k}$ to both transform a vector of landmark locations,  $\mathbf{A}_k(\bx_k),$ and warp an image, $\mathbf{A}_k(I)$. The regression is then performed in the prototype coordinate frame:
\begin{equation}
	\bx_{k+1}' = \bx_{k}' + \mathbf{W}_{k}\bm{\phi}(I',\bx_{k}') + \mathbf{b}_{k}.
	\label{eq:templateRegression}
\end{equation}
The estimated landmark locations in image coordinates are given by the inverse transformation, $\bx_{k+1} = \mathbf{A}_{k}^{-1}(\bx_{k+1}')$.

\begin{figure}[!tb]
\centering	
\includegraphics[width=0.85\linewidth]{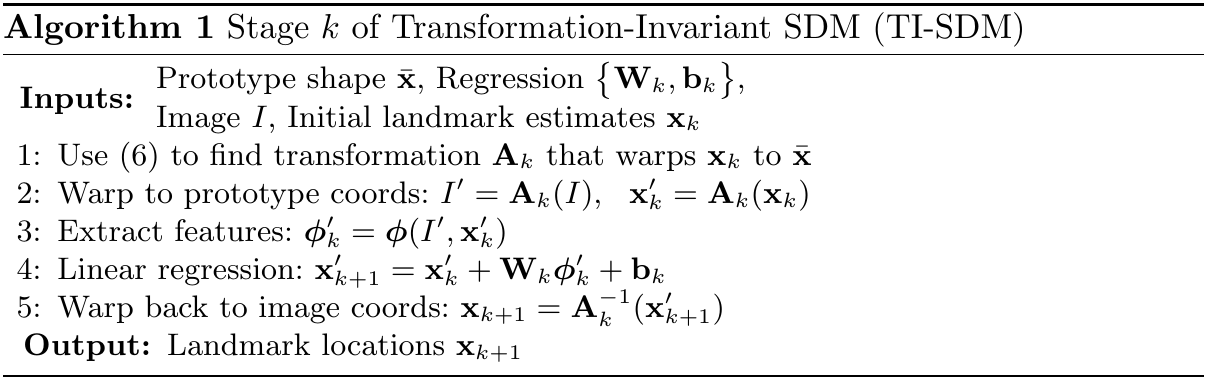}\\[-12pt]
\end{figure}


The resulting algorithm, which we call Transformation-Invariant SDM (TI-SDM), consists of $K$ stages of alignment and regression, illustrated in Fig.~\ref{fig:AlgoDemons}. 
Algorithm~1 
summarizes what happens at each stage of  TI-SDM.

\subsection{Learning deformation constraints}
\label{sec:deformation}

One of the problems associated with using SDM for tracking landmark locations is that it puts no explicit constraint on the regression behavior of neighboring points, which makes it possible for the points to drift apart. This would be a straightforward problem to deal with in an optimization setting by introducing explicit constraints or penalties on the free-form deformation of the landmark points. However, rather than utilizing an optimization procedure, which can be slow, we want to maintain the speed advantages of forward prediction using a regression function. To achieve the effect of constraints within a regression framework, we introduce additional features that allow the regression model to learn to constrain landmark points from drifting.

We introduce a soft constraint in the form of an additional cost term  $||\bx-\bar{\bx}||^2$ in equation \eqref{eq:SDM}:
\begin{equation} 
	f_c(\mathbf{x}_k + \Delta \bx) =
	\norm{\bm{\phi}(I,\mathbf{x}_k + \Delta \mathbf{x})- \hat{\bm{\phi}} }^2 + \lambda \norm{\bx_k + \Delta \mathbf{x} - \bar{\bx}}^2 .
\label{SGDConstraint}
\end{equation}

\begin{figure}[!t]
\begin{minipage}[b]{0.58\textwidth}
  \centering
  \includegraphics[width=\textwidth]{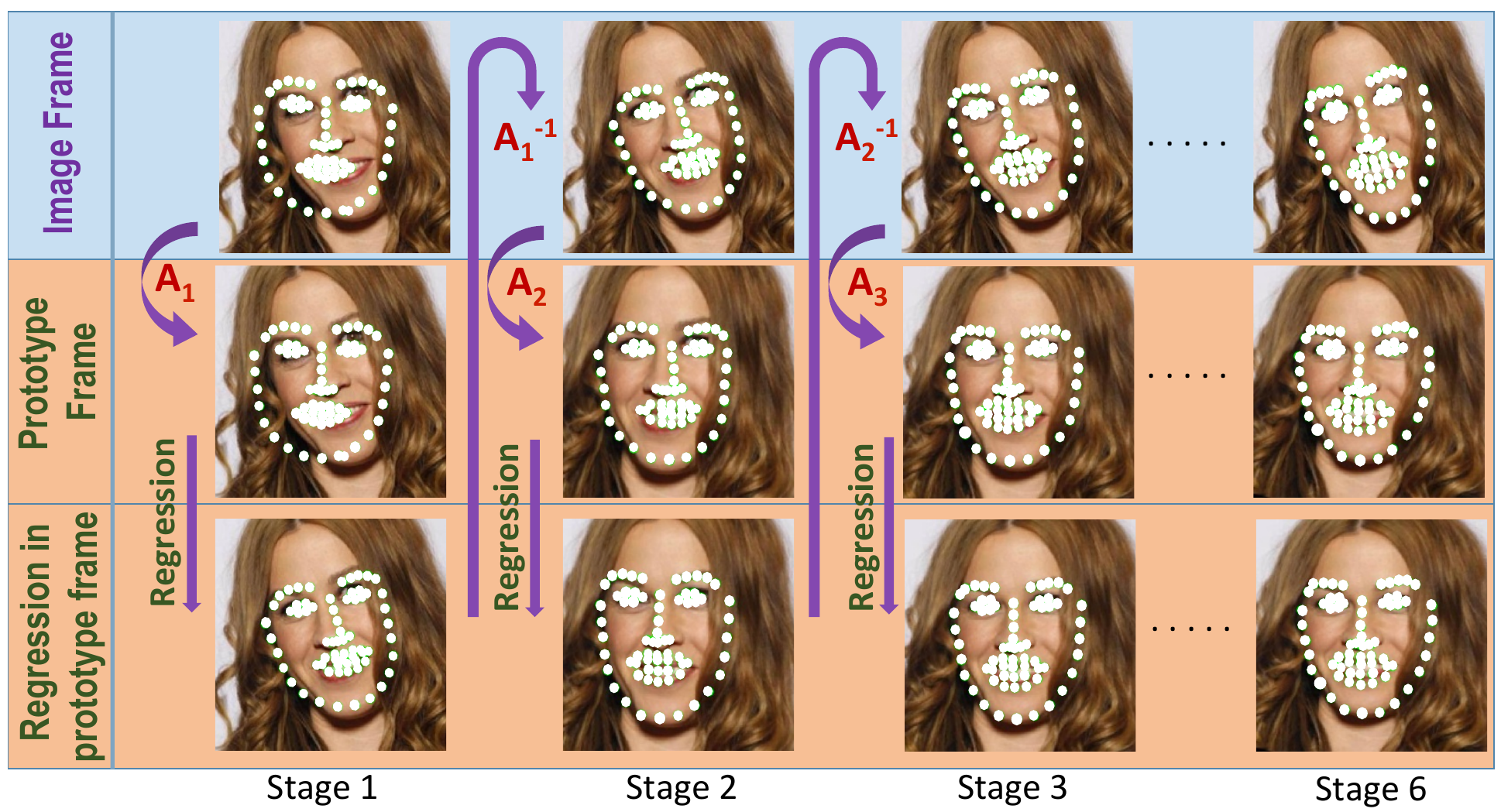}
  \caption{The Transformation-Invariant SDM \mbox{(TI-SDM)} algorithm. See Section~\ref{sec:transformationInvariance} and 
  Algorithm~1 
  for details.}
\label{fig:AlgoDemons}
\end{minipage}
\quad%
\begin{minipage}[b]{0.35\textwidth}
\centering
  \begin{tabular}[t]{ccc} %
  \centering
  {\tiny Initialization} & {\tiny No constraint} & {\tiny With constraint}\\
  \includegraphics[width=0.33\textwidth]{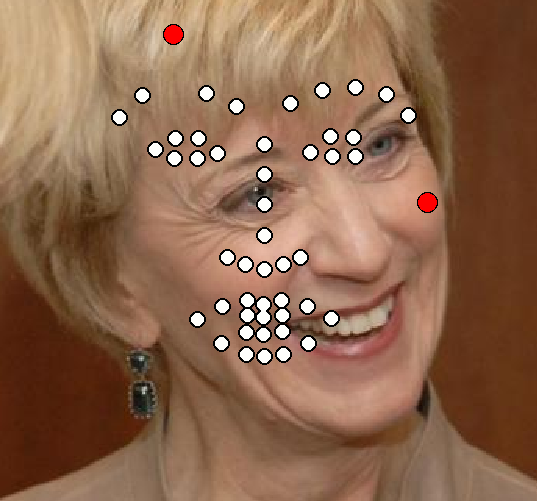}&
	\includegraphics[width=0.33\textwidth]{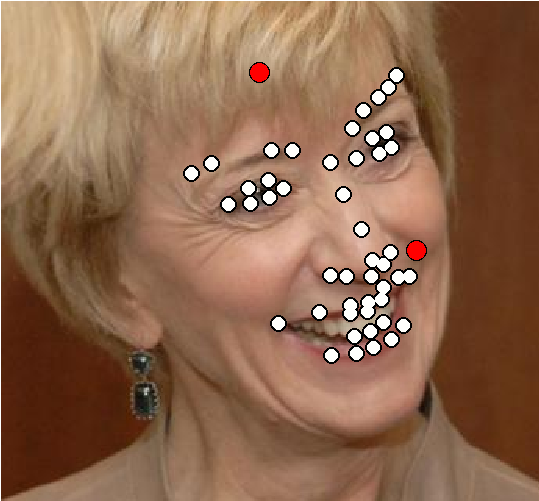}&
	\includegraphics[width=0.33\textwidth]{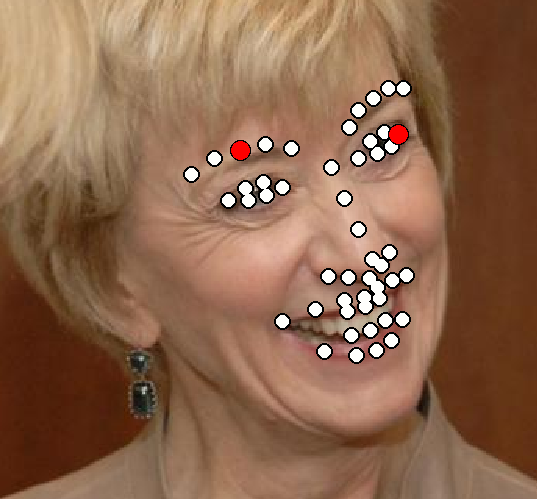}\\[-2pt]
  {\tiny Initialization} & {\tiny No constraint} & {\tiny With constraint}\\
  \includegraphics[width=0.33\textwidth]{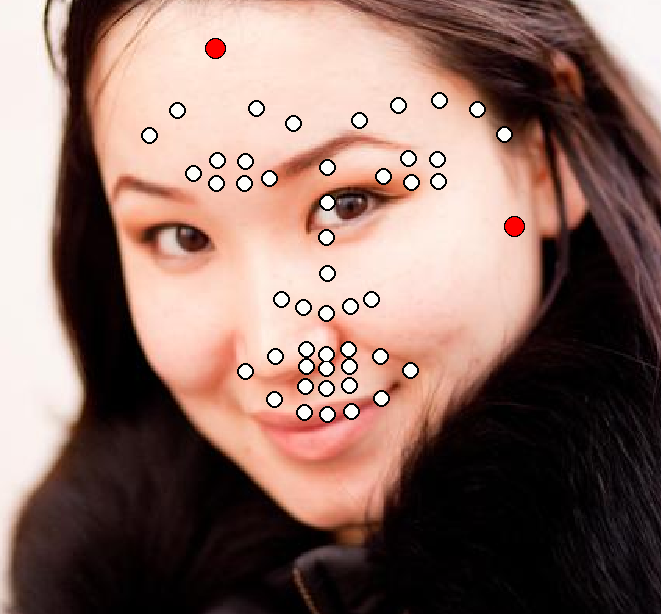}&
	\includegraphics[width=0.33\textwidth]{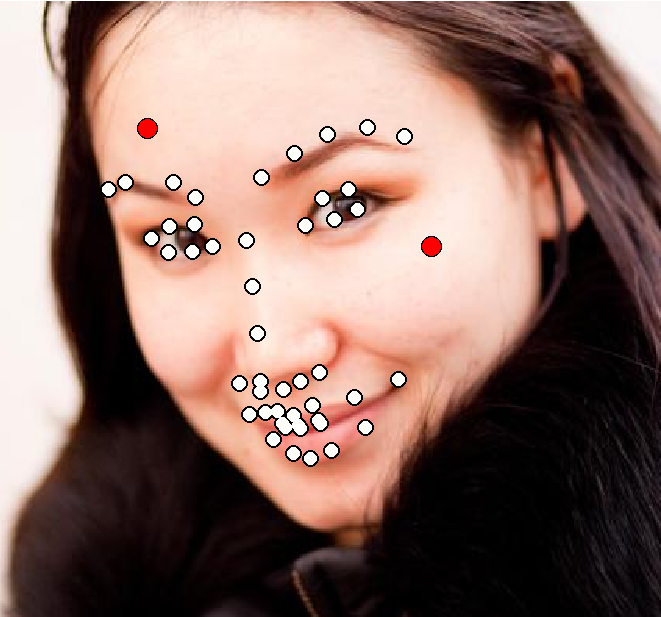}&
	\includegraphics[width=0.33\textwidth]{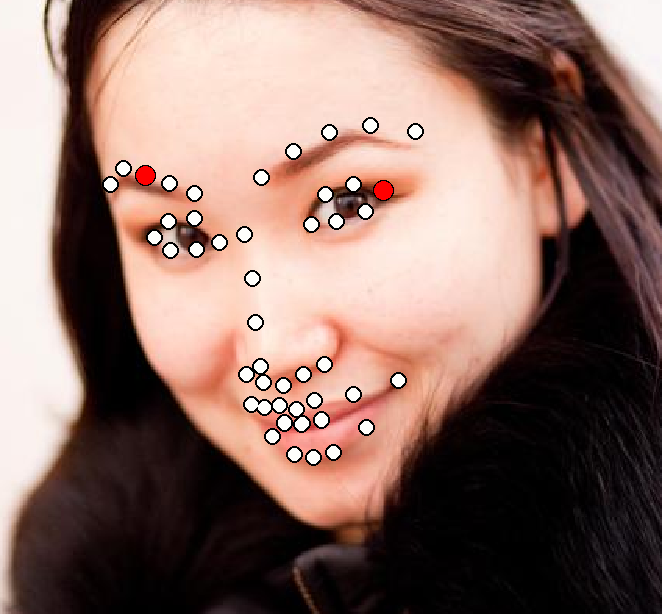}	\\[4pt]	
	\end{tabular}
  \caption{Effect of initial outliers (red) on results {\em without} vs. {\em with} deformation constraint. 
  }
\label{fig:DeformationConstraint}
\end{minipage}
\end{figure}

This enforces a quadratic penalty when the landmark locations drift away from the prototype shape $\bar{\bx}.$ The weight $\lambda$ controls the tradeoff between data and the constraint. The Newton step for this constrained $f$ is given by
\begin{equation}
		\Delta \bx = -2\mathbf{H}^{-1} \begin{pmatrix}
\mathbf{J}_{\bm{\phi}} &  \mathbf{I} 
\end{pmatrix} 
\begin{pmatrix}
\bm{\phi}_k - \hat{\bm{\phi}} \\
\lambda \left(\bx_k-\bar{\bx}\right)
\end{pmatrix},
\label{eq:constrainedNewton}		
\end{equation}
where $\mathbf{H}$ is the Hessian matrix of $f_c$ with respect to $\bx$, and $\mathbf{J}_{\bm{\phi}}$ is the Jacobian of $\bm{\phi}$ with respect to $\bx$. Just as we approximated~\eqref{eq:newton} by~\eqref{eq:regression}, we can approximate this constrained Newton step~\eqref{eq:constrainedNewton} by a linear regression function of a constrained feature vector, $\bm{\phi}^*_k$: 
\begin{equation}
		\label{eq:constrainedRegression}
		\Delta \bx =  \mathbf{W}_k {\bm{\phi}}^*_k+ \mathbf{b}_k, 
		\quad \text{where} \quad
		{\bm{\phi}}^*_k = \begin{pmatrix}
\bm{\phi}_k \\
\lambda \left(\bx_k-\bar{\bx}\right)
\end{pmatrix} .
\end{equation}
As in unconstrained SDM, we can learn the regression coefficients $\mathbf{W}_k $ and bias $\mathbf{b}_k$ using training data. The only difference between the constrained~\eqref{eq:constrainedRegression} and unconstrained~\eqref{eq:regression} regression models is that in the constrained version, we extend the feature vector to include additional features, $\lambda \left(\bx_k-\bar{\bx}\right)$, encoding the deviation of the landmark locations from the prototype landmark locations. In general, during our experiments, the constrained regression learns to move landmark locations towards the mean shape by learning negative values for the associated regression coefficients. The learned coefficients' norms are larger for the initial regression stage of the cascade, but smaller in the later stages, which enforces weaker constraints on deformation as the landmark locations approach convergence.
Note that it would be possible to incorporate $\lambda$ into $\mathbf{W}_k$ and $\bar{\bx}$ into $\mathbf{b}_k$, and just expand the feature vector $\bm{\phi}^*$ with $\bx_k$ rather than $\lambda \left(\bx_k-\bar{\bx}\right)$. However, we choose to keep the difference vector form as in~\eqref{eq:constrainedRegression}, which becomes important for the regularized training described in Section~\ref{sec:training}.

To unify notation, in the rest of this paper we will refer to the expanded feature vector $\bm{\phi}^*$ as simply $\bm{\phi}$. That way, Equations (\ref{eq:regression}\hspace{1pt}--\hspace{1pt}\ref{eq:templateRegression}) and 
Algorithm~1 
apply to the constrained model without modification. In Fig.~\ref{fig:DeformationConstraint}, we analyze the effect of the deformation constraint. See Section~\ref{sec:Exp} for details.

\subsection{Mixture-of-experts regression}
\label{sec:mixture}

The transformation invariance step described in Section~\ref{sec:transformationInvariance} lets our model learn regression functions that are invariant to affine transformations of the faces. Still, the remaining variations in the data (e.g., due to out-of-plane rotations and facial expressions) are large enough that it is challenging for a single regression function to accurately align all faces. In particular, the training set in our experiments includes many more frontal faces with mild facial expressions than faces with large out-of-plane rotations or extreme expressions. Thus, the prototype (mean) face is close to a frontal face with neutral expression, and the regression function tends to work less well for more extreme poses and expressions. 


We propose to use a mixture-of-experts regression model, in which each expert is a regression function that is specialized for a different subset of the possible poses and expressions. Each expert's subset is determined by the expert's prototype shape. We construct $L$ prototype shapes, $\{\bar{\bx}^l\}_{l=1, \ldots, L}$, such that the set of ground-truth landmark locations $\hat{\mathbf{x}}_n$ of each of the $N$ faces in the dataset is well aligned with one of the prototype shapes. We write the determination of the prototype shapes as an optimization problem:
\begin{equation} 
		\{\bar{\bx}^l\}_{l=1, \ldots, L} = \argmin_{\{\dot{\bx}^l\}_{l=1, \ldots, L}} \; \sum_{n=1}^{N} \; \min_{\substack{\mathbf{A} \in \mathcal A,\\ l \in \{1, \ldots, L\}}} \! \norm{\mathbf{A}(\hat{\bx}_n) - \dot{\bx}^l}^2,	
		\label{eq:clustering} 
\end{equation}
where each $\dot{\bx}^l$ is a $2p \times 1$ vector representing a possible prototype face shape (i.e., the locations of $p$ landmarks). 
If the class of transformations, $\mathcal A$, only contains the identity transformation, then this problem reduces to Euclidean clustering of training samples based on landmark locations (see Fig.~\ref{fig:MixtureAndClusters}a). 

When $\mathcal A$ is the class of affine transformations, we call this affine-invariant clustering. In this case, \eqref{eq:clustering} is a homogenous optimization problem in which additional constraints on the prototype shapes or the transformations are necessary to avoid the zero solution (which assigns zero to all of the transformations and prototype shapes). Moreover, the objective function is non-convex due to the joint optimization of the shapes and the assignment of training samples to shapes. We decouple this problem into two convex sub-problems, which we solve iteratively. The first sub-problem assigns every training face image $n$ to one of the prototype shapes via the equation
\begin{equation} 
		l_n = \argmin_{l} \left[\min_{\mathbf{A} \in \mathcal A} \norm{\mathbf{A}(\hat{\bx}_n) - \bar{\bx}^l}^2 \right]
		\label{eq:assignment} 
\end{equation}
assuming that the prototype shapes $\bar{\bx}^l$ are fixed. This problem can be solved independently for each training face: The optimal assignment is the prototype to which the face's ground-truth landmark locations can be affine-aligned with minimum alignment error. The second sub-problem solves for the prototype shapes. Each prototype shape consists of the landmark locations that minimize the sum of the squared affine alignment errors of the ground-truth locations $\hat{\bx}_n$ of the training faces that were assigned to that prototype shape:
\begin{equation} 
		\bar{\bx}^l = \argmin_{\dot{\bx}^l}  \sum_{n \text{ s.t. } l_n = l}   \min_{\mathbf{A} \in \mathcal A} \norm{\mathbf{A}(\hat{\bx}_n) - \dot{\bx}^l}^2 
		 \quad s.t. 	\quad   \mathbf{C} \, \dot{\bx}^l = \mathbf{m}, \label{eq:constraint}
\end{equation}
where to avoid degeneracy, the matrix $\mathbf{C}$ and vector $\mathbf{m}$ impose linear constraints on the prototype shape such that the mean location of the 5 landmark points of the left eyebrow is fixed, as are the mean location of the 5 right eyebrow points and the mean vertical location of the 16 mouth points. This optimization problem is quadratic with linear constraints, and the optimal solution is computed by solving a linear system. The two optimization sub-problems are alternately solved until the assignments do not change. In our experiments, 20--30 iterations suffice for convergence.

In Fig.~\ref{fig:MixtureAndClusters} (right), we compare Euclidean clustering (a) with the proposed affine-invariant clustering (b). Euclidean clustering only accounts for the pose variations in the dataset. However, some of the out-of-plane poses can be approximately aligned to each other with an affine alignment, enabling the affine-invariant clustering to account for variations in both pose and facial expressions.

Each expert $E^l$ corresponds to one of the $L$ prototype shapes. At each stage of the regression cascade, we learn a separate regression for each expert. Hence,
in addition to its prototype shape $\{\bar{\bx}^l\}$, each regression expert $E^l$ has a regression function $\left\{ \mathbf{W}^l_k, \bb^l_k\right\}$ for each of the $K$ levels of the cascade: 
\begin{equation}
	E^l = \left\{ \bar{\bx}^l, \ \ \left\{ \mathbf{W}^l_k, \bb^l_k\right\}_{k=1, \ldots, K} \right \}.
\end{equation}
At each stage, $k$, of the cascade, each expert $E^l$ performs 
Algorithm~1 
using prototype $\bar{\bx}^l$ and regression function $\bigl \{ \mathbf{W}^l_k, \bb^l_k \bigr \}$:
\begin{equation}
\bx^l_{k+1} = 
\text{Algorithm\,1} 
\! \left( \bar{\bx}^l, \bigl \{ \mathbf{W}^l_k, \bb^l_k \bigr \}, I, \bx_k \right).
\end{equation}


The gating function for each regression expert $E^l$ is a soft assignment $\alpha^l (\bx_k)$ given by the softmax transformation of the transformation invariance error $\epsilon^l (\bx_k)$ between the starting landmark locations $\bx_k$ and each prototype shape~$\bar{\bx}^l.$ The soft assignments are computed using 
\begin{equation} 
\label{eq:gating}
	\alpha^l(\bx) = \frac{e^{-\epsilon^l(\bx)}}{\sum_{l=1}^L e^{-\epsilon^l(\bx)}}, \quad
	\text{where} \qquad  
	\epsilon^l(\bx) = \min_{\mathbf{A} \in \mathcal A} \norm{\mathbf{A}(\bx) - \bar{\bx}^l}^2.
\end{equation}
Here, as in~\eqref{eq:affine}, $\mathcal{A}$ denotes the set of all affine transformations.
A high score $\alpha^l(\bx_k)$ indicates that the current estimate $\bx_k$ is close to the prototype shape of the $l$th expert, and hence the regression results obtained from $E^l$ would be given a high weight. In Fig.~\ref{fig:MixtureAndClusters} (left), we show the assignment weights of two faces to experts in the model. 

At each stage, $k$, of the cascade, our alignment algorithm applies every expert's regression function to the starting estimate of landmark locations $\bx_k$, then averages the outputs according to the gating function~$\alpha^l(\bx_{k})$ to obtain the updated estimate of landmark locations, $\bx_{k+1}$:
\begin{equation}
	\bx_{k+1} = \sum_{l=1}^L \alpha^l(\bx_{k}) \, \bx^l_{k+1}.
\end{equation}
Algorithm~2 
summarizes our alignment method, which we call {\bf M}ixture of 
\linebreak 
{\bf I}nvariant E{\bf x}perts (MIX).

Note that our mixture-of-experts model is quite different from multi-view models \cite{romdhani1999multiview,cootes2002viewAAM,Asthana2011ICCV3DposeNorm,vcech2016multi}, which explicitly model and predict the head pose (e.g., by learning a different deformable model for each of several specific pose ranges). In contrast, MIX is a discriminative mixture model that discovers a data-depen\-dent partitioning of the shape space (see Fig.~\ref{fig:MixtureAndClusters}b) based on facial expressions and other affine-invariant shape variations (including affine-invariant variations due to pose), and learns a different optimization for each partition.  

\begin{figure}[!tb]
\centering	
\includegraphics[width=0.85\linewidth]{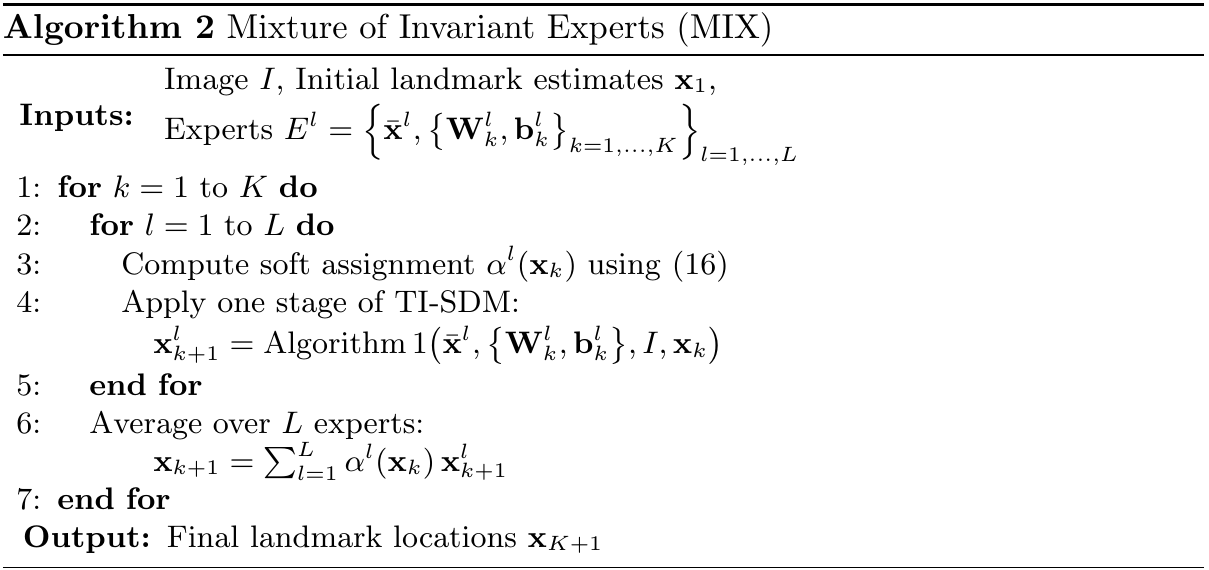}\\[-12pt]
\end{figure}

%

\subsection{Training the experts model}
\label{sec:training}


To learn the regression experts $E^l,$ the $N$ face images in the training data are augmented by repeating every training image $M$ times, each time perturbing the ground-truth landmark locations by a different random displacement. For each image $I_i$ in this augmented training set $(i = 1, \ldots, M\!N),$ with ground-truth landmark locations $\hat{\bx}_i$, we displace the landmarks by a random displacement $\Delta \hat{\bx}_i$. 
For every expert $E^l$, we use~\eqref{eq:gating} to compute the soft assignment $\alpha^l_i$ of the $i$th sample's perturbed landmark locations to the prototype shape $\bar{\bx}^l$:
\begin{equation}
\alpha^l_i = \alpha^l \left( \hat{\bx}_i + \Delta \hat{\bx}_i \right).
\end{equation}
While computing this soft assignment, let $\mathbf{A}_i^l$ denote the global (affine) transformation from~\eqref{eq:gating} that best aligns the $i$th sample's perturbed landmark locations to prototype shape $\bar{\bx}^l$. Use $\mathbf{A}_i^l$ to transform the ground-truth landmark locations and displacement vectors into the prototype coordinate frame of expert $E^l$:
\begin{equation}
\hat{\bx}_i^l = \mathbf{A}_i^l(\hat{\bx}_i), \qquad \Delta \hat{\bx}_i^l = \mathbf{A}_i^l( \Delta \hat{\bx}_i).
\end{equation}

\begin{figure*}[!tb]
  \centering	
  \includegraphics[width=0.32\linewidth, trim={0 0 0 23pt},clip]{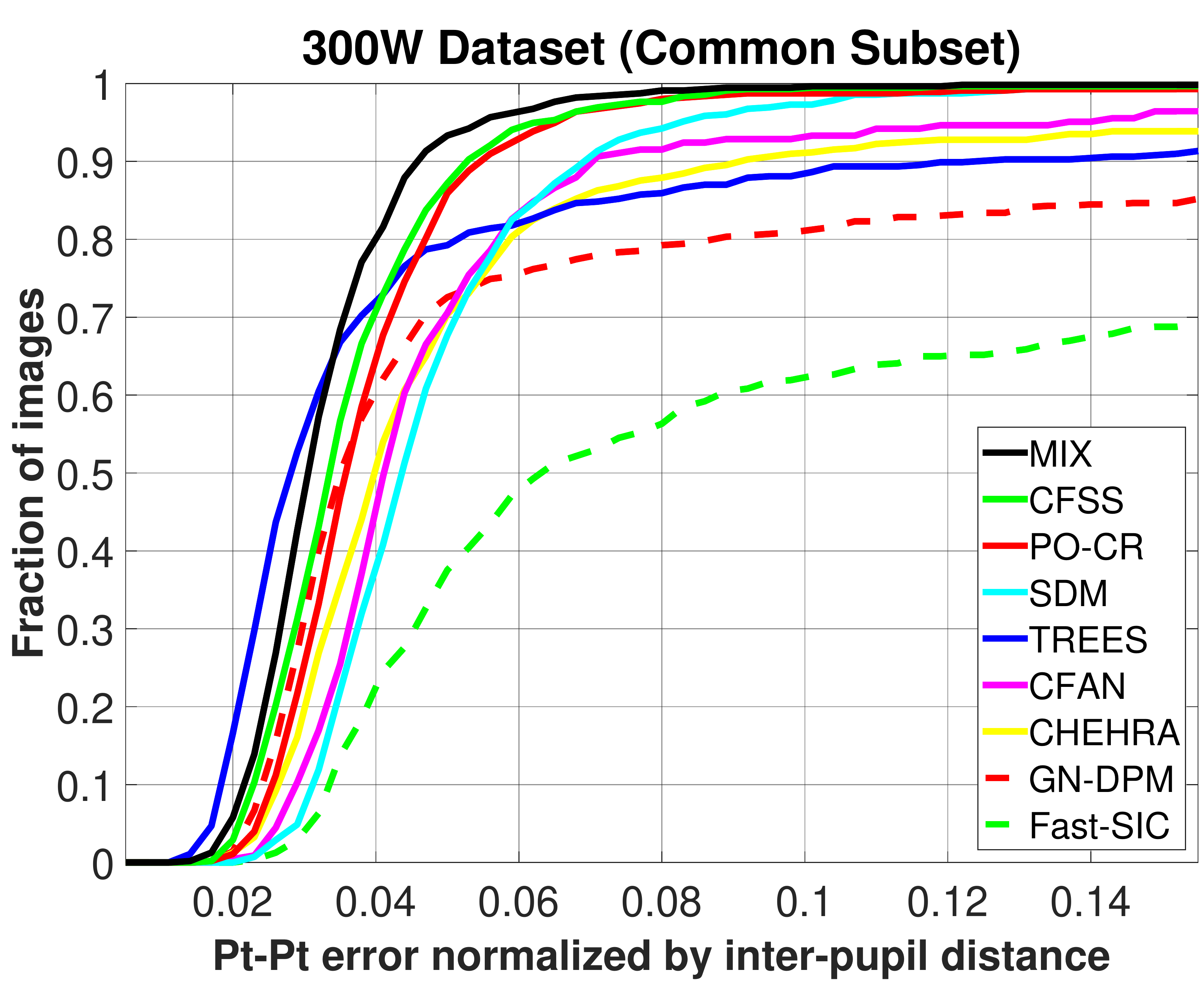}
  \includegraphics[width=0.32\linewidth, trim={0 0 0 23pt},clip]{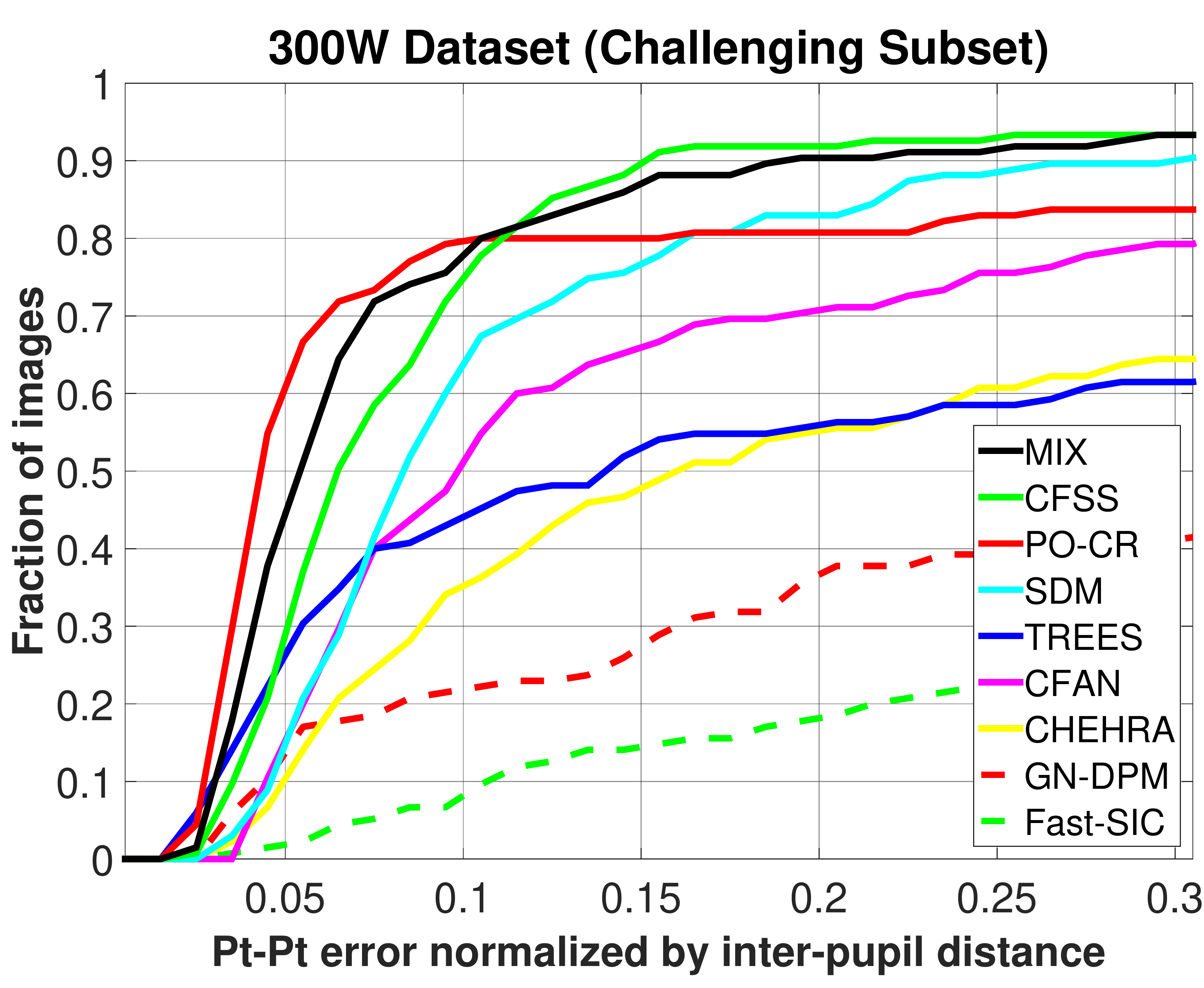} 
  \includegraphics[width=0.32\linewidth, trim={0 0 0 23pt},clip]{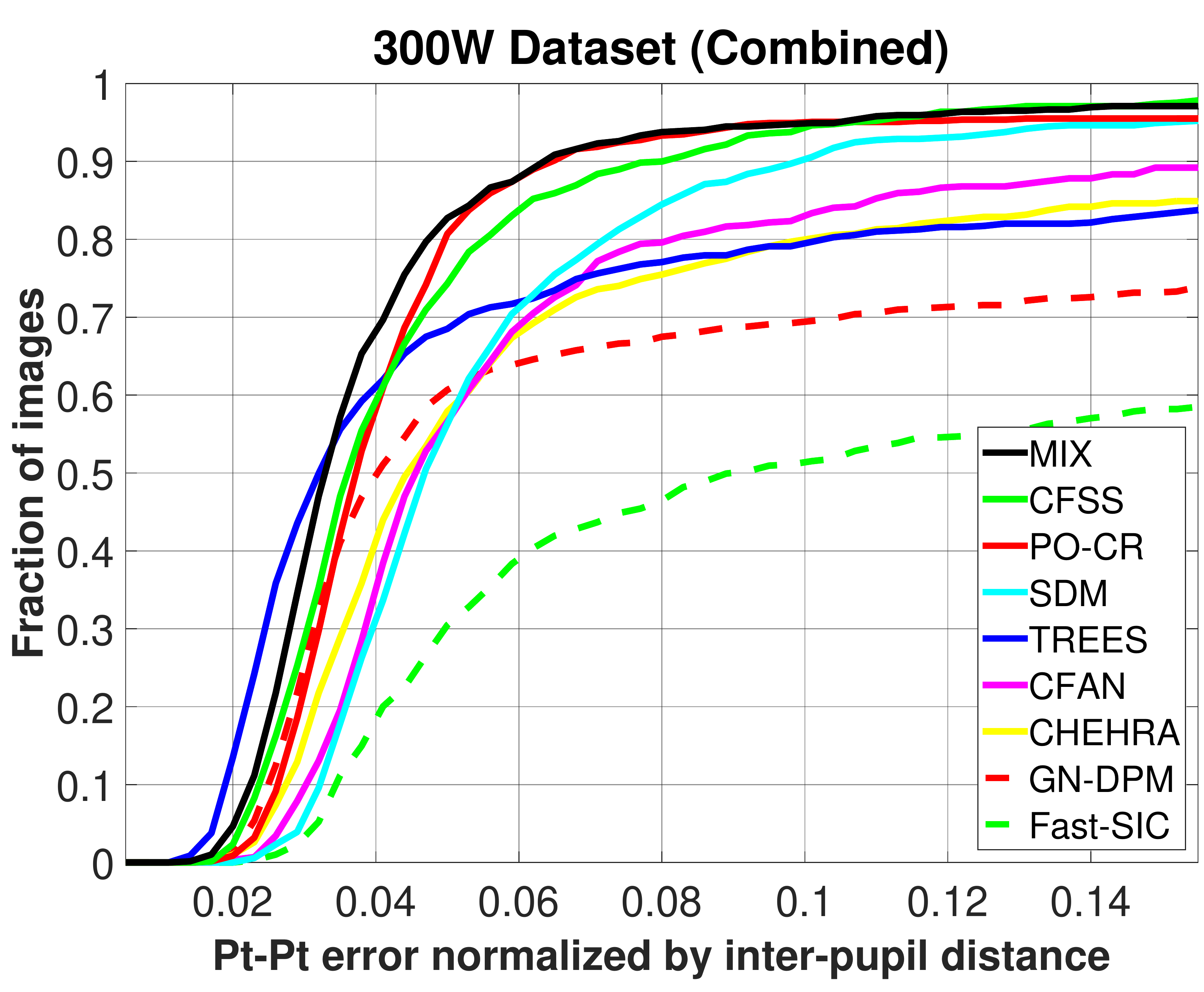}\\[-6pt]
  \caption{Comparison of MIX with other state-of-the-art methods on the 300W dataset.}
\label{fig:Curves300W}
\end{figure*}

\begin{figure*}[!tb]
  \centering	
  \includegraphics[width=0.32\linewidth, trim={0 0 0 23pt},clip]{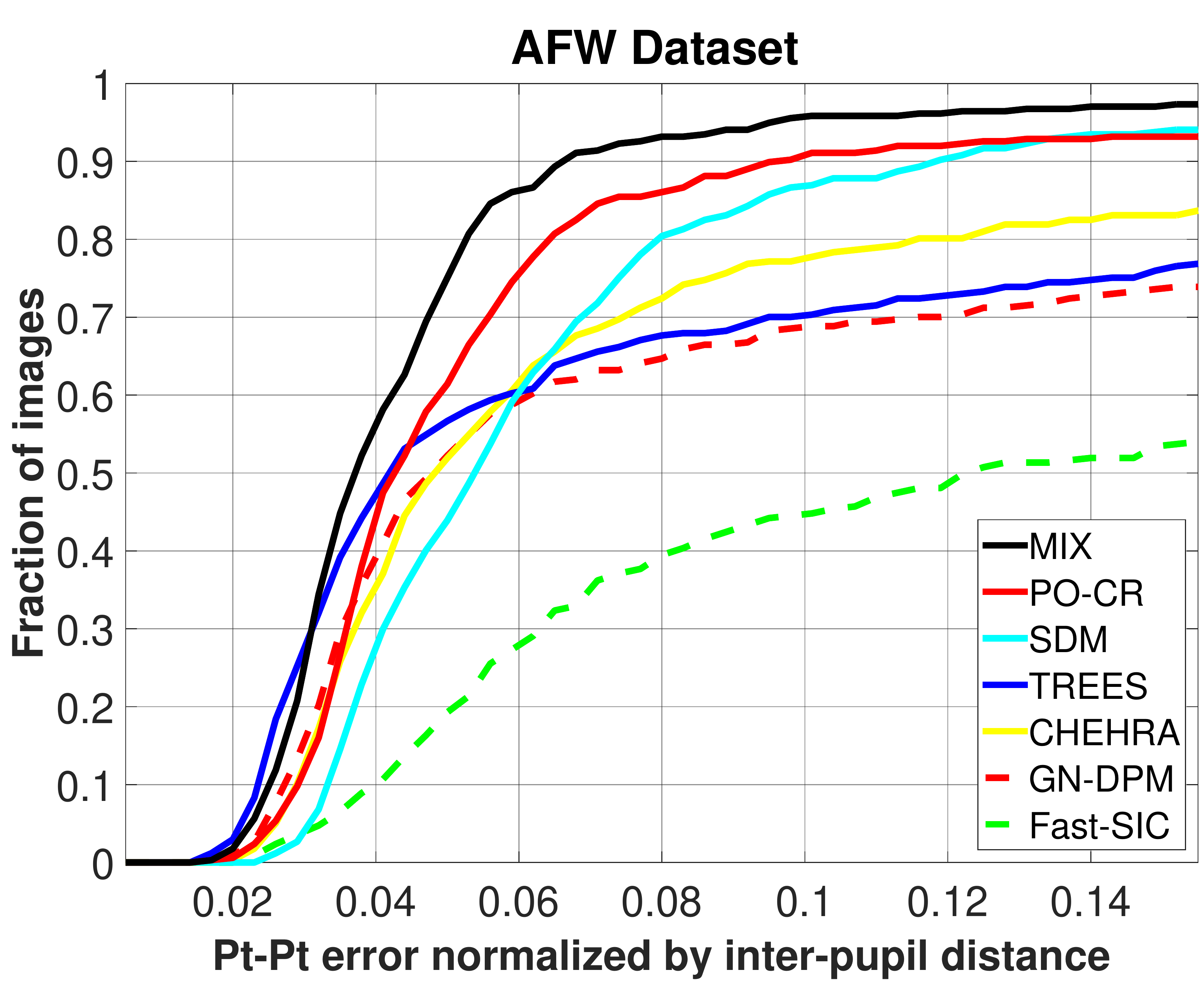}
  \qquad \qquad
  \includegraphics[width=0.32\linewidth, trim={0 0 0 23pt},clip]{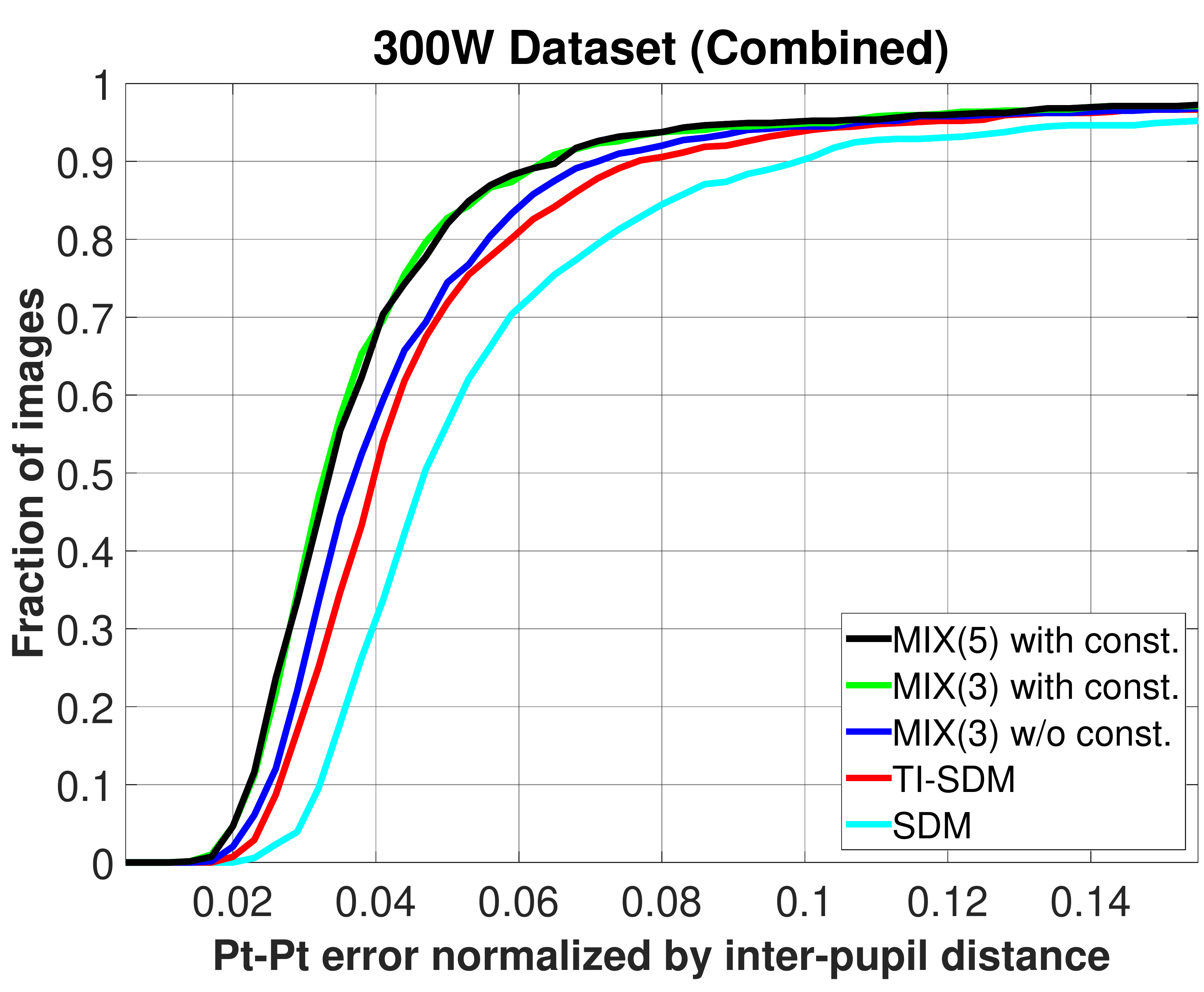}\\[-6pt]
\caption{{\em Left:} Comparison of MIX with other state-of-the-art methods on the AFW dataset. CFAN and CFSS are not compared, because both included AFW in their training set.  {\em Right:}  Comparing variations of proposed method on 300W (Combined).}
\label{fig:CurvesAFWandvariations}
\end{figure*}

The first regression function ($k=1$) is then learned by minimizing a Tikhonov regularized L2-loss function:
\begin{equation}
	\{ \mathbf{W}^l_k, \mathbf{b}^l_k \} =
 \argmin_{\mathbf{W}, \mathbf{b}} \sum_{i=1}^{M\!N} \alpha^l_i \| \Delta \hat{\bx}_i^l \!-\! \mathbf{W} \bm{\phi}(I_i,\hat{\bx}_i^l \!-\! \Delta \hat{\bx}_i^l) \!-\! \mathbf{b}\|^2 
	+ \gamma \left[\|\mathbf{W}\|_{F}^2 +  \|\mathbf{b}\|_{F}^2\right] .
\end{equation}
For each $l$ and $k$, the regularizer weight $\gamma$ is selected via grid search in log space using 2-fold cross validation. 

For training the later regressions $\{\mathbf{W}_k, \mathbf{b}_k\}_{k=2, \ldots, K},$ rather than using a random perturbation, the target $\Delta \hat{\bx}_i$ is the residual of the previous stages of the cascade. In training, the regression function may diverge for a few samples, producing large residuals. To avoid fitting these outliers, at each stage $k$, we remove $5\%$ of the samples with the largest residuals from the training set. We choose the number of regression stages $K$ by training until the cross-validation error cannot be reduced further.

The training samples are generated by randomly perturbing the ground-truth facial landmark locations along the major deformation directions of the training set, which are determined via principal component analysis. In addition, we apply random rotation, translation, and anisotropic scaling to the landmark locations, and add i.i.d. Gaussian noise. After learning the cascade model for this training set (usually $K =$ 3\hspace{1pt}--\hspace{1pt}4 stages), we learn a second cascade model using a training set consisting of only small amount of i.i.d. Gaussian noise, and append this model to the original model. The second model has 1--2 stages and improves fine alignment. 

During testing, the initial landmark locations for regression are given by the mean landmark locations from all of the training data, translated and scaled to fit the given face detector bounding box.


\section{Experiments} \label{sec:Exp}


\begin{table}[t]
\centering
\caption{\label{tab:naucResults}Numerical comparison of all tested methods on the 300W (Combined) dataset.}
\begin{scriptsize}
\begin{tabular}{|c|c|c|c|c|c|}
\hline
Method & NAUC$_{0.1}$ & NAUC$_{0.2}$ & NAUC$_{0.3}$ & NAUC$_{0.4}$ & NAUC$_{0.5}$\\ \hline
MIX & {\bf 0.5945} & {\bf 0.7810} & {\bf 0.8482} & {\bf 0.8828} & {\bf 0.9043}\\ \hline
CFSS~\cite{Zhu2015cvprCFSS} & 0.5528 & 0.7613 & 0.8354 & 0.8733 & 0.8967\\ \hline
PO-CR~\cite{tzimiropoulos2015POCR} & 0.5610 & 0.7568 & 0.8242 & 0.8588 & 0.8798\\ \hline
SDM~\cite{xiong2013supervised} & 0.4475  & 0.6957 & 0.7880 & 0.8362 & 0.8662\\ \hline
TREES~\cite{kazemi2014one} & 0.5187  & 0.6746 & 0.7406 & 0.7792 & 0.8063\\ \hline
CFAN~\cite{Zhang2014eccvCFAN} & 0.4357  & 0.6594 & 0.7487 & 0.7985 & 0.8317\\ \hline
CHEHRA~\cite{asthana2014incremental} & 0.4376  & 0.6390 & 0.7217 & 0.7703 & 0.8038\\ \hline
GN-DPM~\cite{tzimiropoulos2014gauss} & 0.4274  & 0.5796 & 0.6531 & 0.7000 & 0.7354\\ \hline
Fast-SIC~\cite{tzimiropolous2013aam} & 0.2490  & 0.4122 & 0.4984 & 0.5644 & 0.6179\\ \hline
\end{tabular}
\end{scriptsize}
\end{table}

In the first experiment, we compare our proposed algorithm (MIX) to eight state-of-the-art algorithms: CFAN~\cite{Zhang2014eccvCFAN}, TREES~\cite{kazemi2014one}, CFSS~\cite{Zhu2015cvprCFSS}, SDM~\cite{xiong2013supervised}, CHEHRA \cite{asthana2014incremental}, GN-DPM (using SIFT features)~\cite{tzimiropoulos2014gauss}, \mbox{Fast-SIC}~\cite{tzimiropolous2013aam}, and PO-CR~\cite{tzimiropoulos2015POCR}. We evaluate performance on the 300W  \cite{sagonas2013CVPRlabels,sagonas2013ICCV300w} and AFW~\cite{zhu2012face} datasets.

We train our MIX algorithm on the training sets of two standard datasets: LFPW~\cite{belhumeur2011localizing} ($811$ training faces) and Helen~\cite{le2012interactive} ($2000$ training faces). We augment the training data by horizontally flipping each image, yielding $N =$ 5,622 training images. From each image, we sample $M=15$ training initializations (see Section~\ref{sec:training}). We use 3 experts ($L = 3$), because using more ($L=5$) did not significantly improve performance (see the second experiment, below).  For SDM, we use our own implementation, trained on the same training data as MIX. 

For the other seven methods, we use their authors' publicly available code. Note that the training set for our algorithm is a smaller subset of the training sets used by CFSS~\cite{Zhu2015cvprCFSS} and CFAN~\cite{Zhang2014eccvCFAN}. Both CFSS and CFAN include  AFW ($337$ faces) in the training set, but we do not, opting instead to test on AFW.

We test all methods on 300W using the same test set as~\cite{Zhu2015cvprCFSS}, which comprises the test sets of LFPW~\cite{belhumeur2011localizing} ($224$ test faces) and Helen~\cite{le2012interactive} ($330$ test faces)\footnote{The CFAN~\cite{Zhang2014eccvCFAN} algorithm included the $330$ test faces from Helen in its training data. Thus when testing CFAN, we had to omit these $330$ faces from the 300W test set.} 
as well as the IBUG dataset ($135$ test faces). For all test images, we used the bounding box initializations provided on the 300W website (face detector bounding boxes). 
To compute errors of results, for all datasets we used the ground-truth locations of 49 landmarks from \cite{sagonas2013CVPRlabels,sagonas2013ICCV300w}. As in~\cite{Zhu2015cvprCFSS}, the 300W {\em common subset} contains the test samples from LFPW (224) and HELEN (330), the {\em challenging subset} is IBUG (135), and {\em combined} refers to all 689 test images. Fig.~\ref{fig:Curves300W} plots the cumulative distribution of the fraction of images, as a function of error normalized by the inter-pupil distance.

Table~\ref{tab:naucResults} presents a numerical comparison of our MIX algorithm with the previous eight methods on the entire (combined) 300W test set. Rather than measuring mean error, which is extremely sensitive to outliers with large alignment error~\cite{YangJLR15arxiv}, we instead use a normalized variation of the AUC$_\alpha$ error metric proposed by~\cite{YangJLR15arxiv}. The error metric we use, Normalized AUC$_\alpha$ (NAUC$_\alpha$), measures the area under each cumulative distribution curve (the curves in Fig.~\ref{fig:Curves300W}) up to a threshold normalized error value $\alpha$, then divides by $\alpha$ (the maximum possible area for that threshold). The resulting NAUC$_\alpha$ error measure, indexed by~$\alpha$, is always between 0 and 1 (where 1 is a perfect score):
$
\text{NAUC}_{\alpha} = \frac{1}{\alpha} \int_0^{\alpha}f(e)de ,
$
where $e$ is the normalized error, $f(e)$ is the cumulative error distribution function, and $\alpha$ is the upper bound that is used to calculate the definite integral. 

The results in Fig.~\ref{fig:Curves300W} and Table~\ref{tab:naucResults} show that our method outperforms all of the other recent methods on 300W. Note that for the next best method, CFSS, we used the code provided by the authors (the more accurate, but slower, version described in~\cite{Zhu2015cvprCFSS}), which is not practical for real-time use: the CFSS code required 1.7 s per face on our machine.

The evaluation results on the AFW dataset (Fig.~\ref{fig:CurvesAFWandvariations}, left) show a similar trend, in which our algorithm outperforms the other methods. CFAN and CFSS are not compared on AFW, because both included AFW in their training set.

\begin{figure*}[!tb]
  \centering
\begin{tabular}[t]{cccccccc}
\adjincludegraphics[width=0.115\textwidth,height=.1125\textwidth,trim=0 {0.1\height} 0 0,clip]{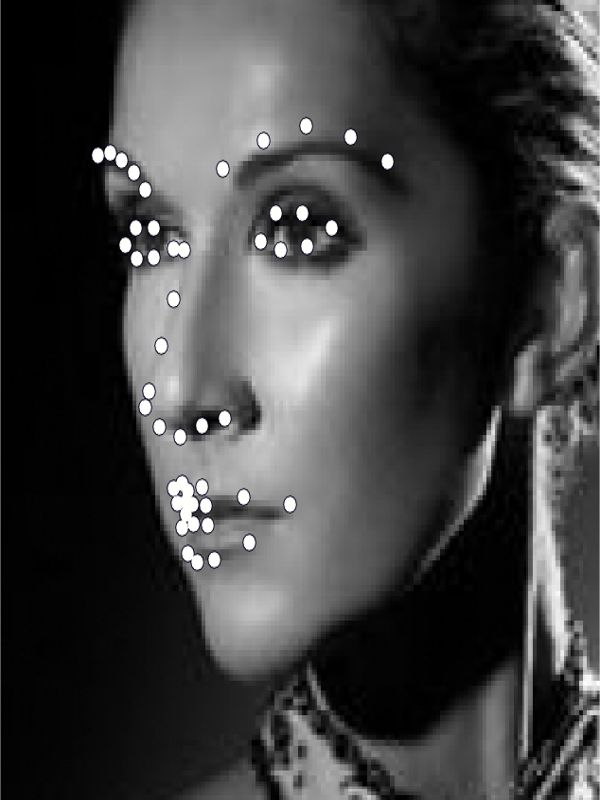}& 
\adjincludegraphics[width=0.115\textwidth,height=.1125\textwidth,trim=0 {0.1\height} 0 0,clip]{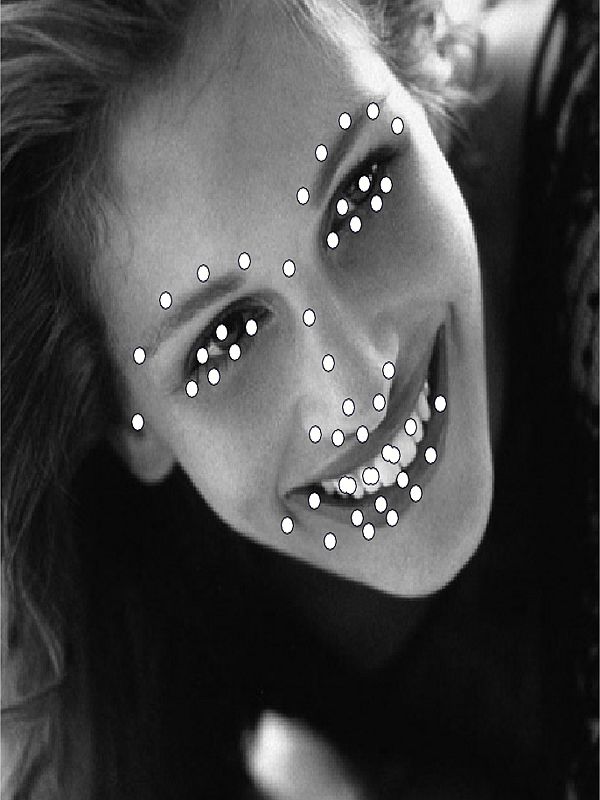}& 
\adjincludegraphics[width=0.13\textwidth,height=.1125\textwidth,trim=0 {0.1\height} 0 0,clip]{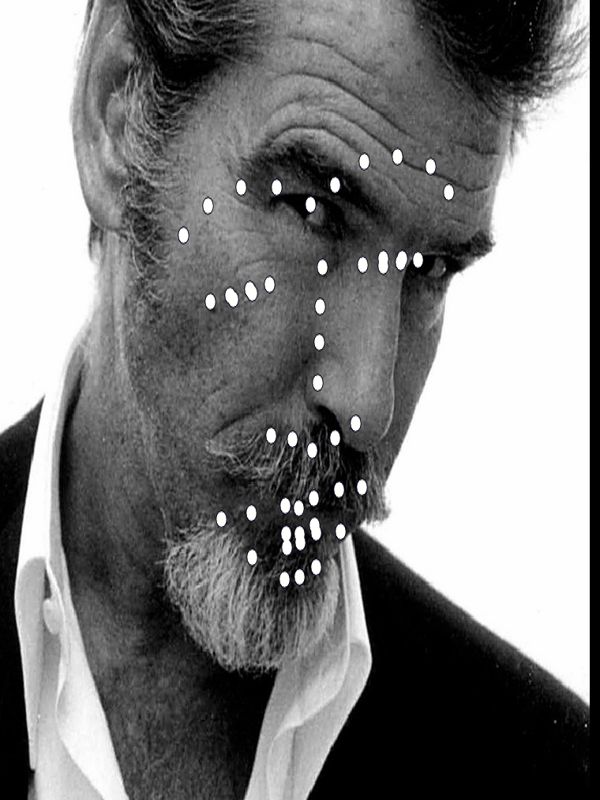}& 
\adjincludegraphics[width=0.115\textwidth,height=.1125\textwidth,trim=0 {0.1\height} 0 0,clip]{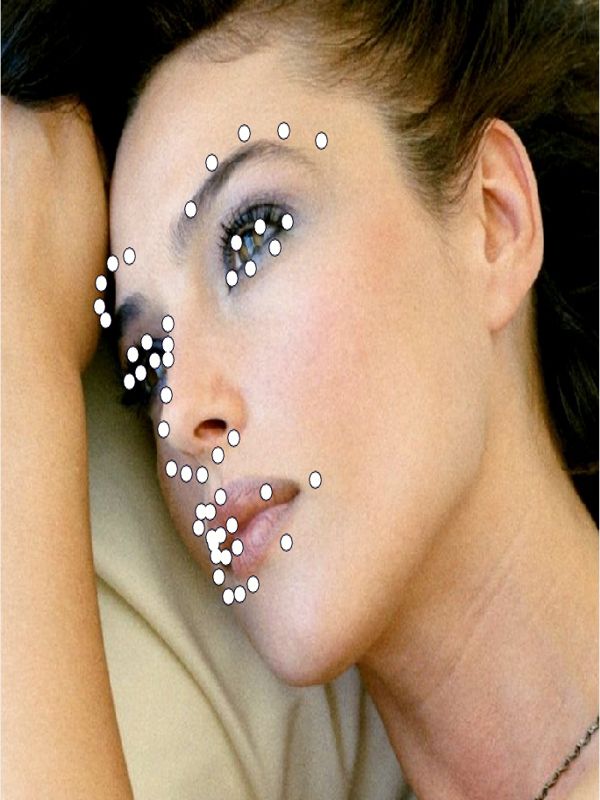}& 
\adjincludegraphics[width=0.115\textwidth,height=.1125\textwidth,trim=0 {0.1\height} 0 0,clip]{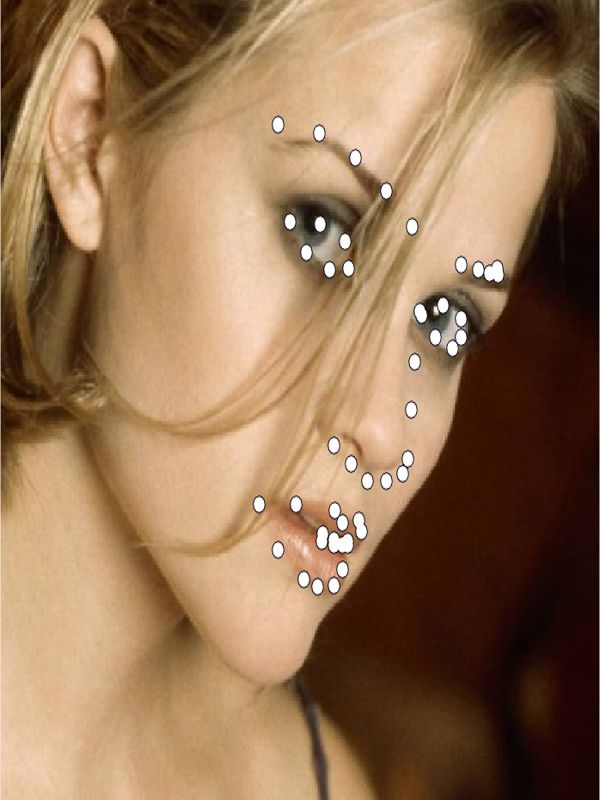}& 
\adjincludegraphics[width=0.115\textwidth,height=.1125\textwidth,trim=0 {0.1\height} 0 0,clip]{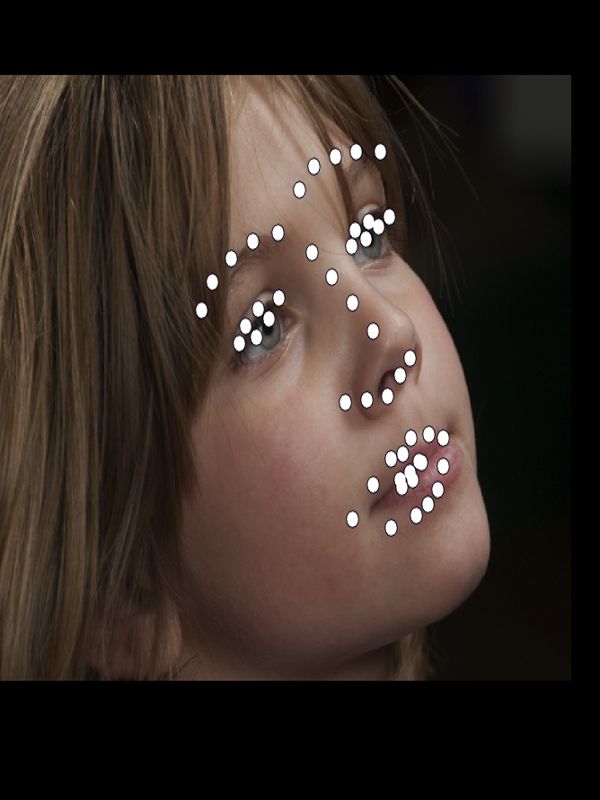}& 
\adjincludegraphics[width=0.115\textwidth,height=.1125\textwidth,trim=0 {0.1\height} 0 0,clip]{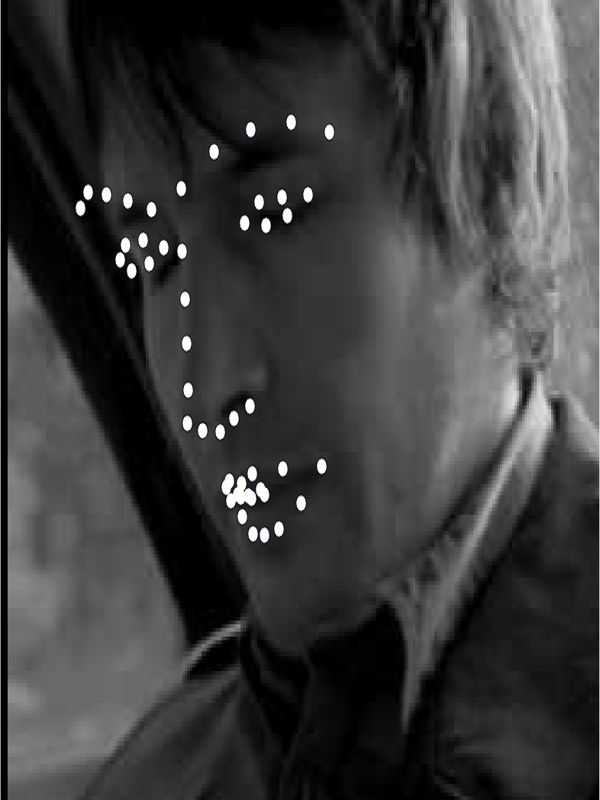}& 
\adjincludegraphics[width=0.115\textwidth,height=.1125\textwidth,trim=0 {0.1\height} 0 0,clip]{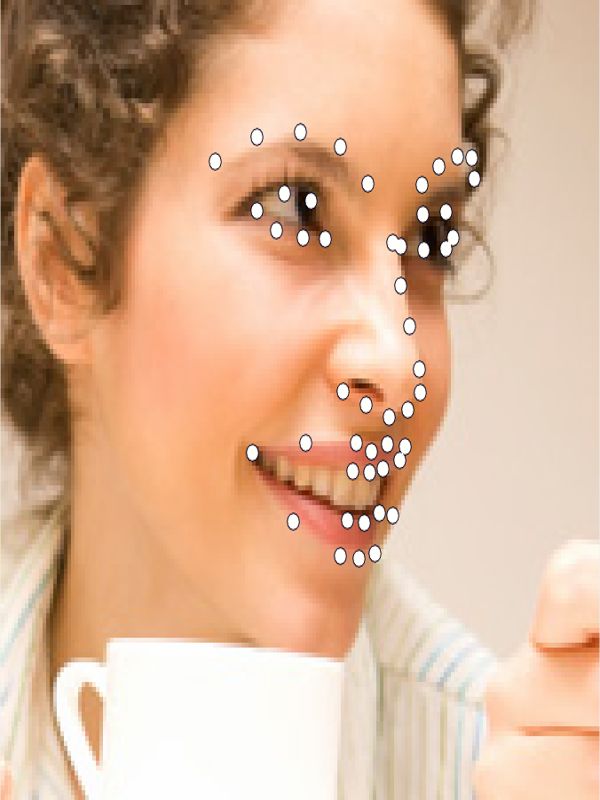}\\ 
\adjincludegraphics[width=0.115\textwidth,height=.1125\textwidth,trim=0 {0.1\height} 0 0,clip]{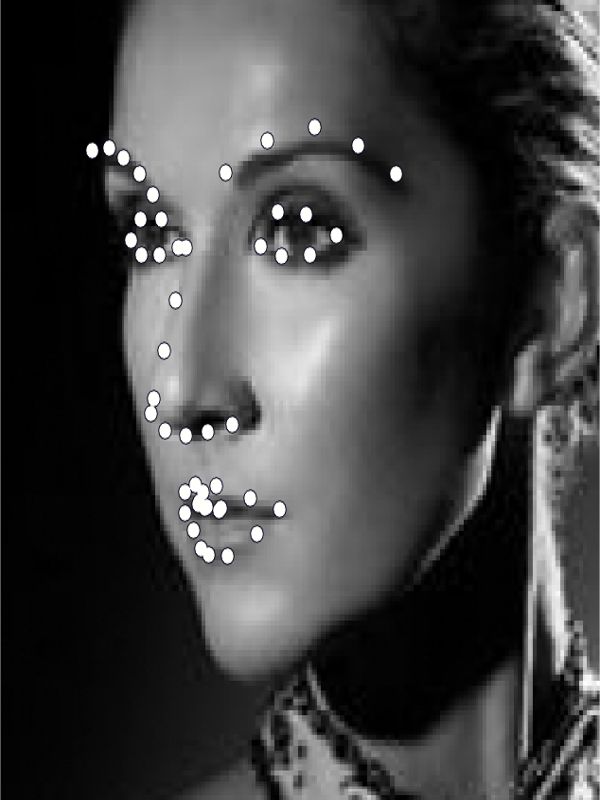}& 
\adjincludegraphics[width=0.115\textwidth,height=.1125\textwidth,trim=0 {0.1\height} 0 0,clip]{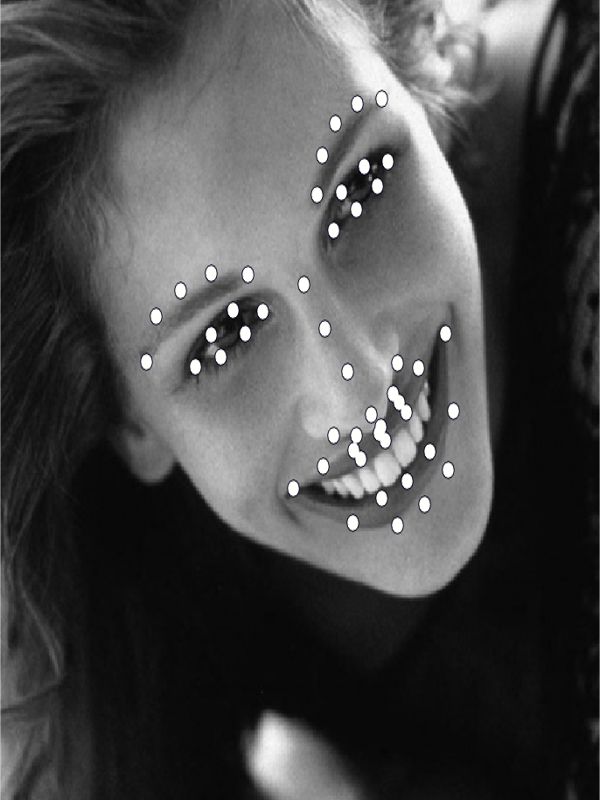}& 
\adjincludegraphics[width=0.13\textwidth,height=.1125\textwidth,trim=0 {0.1\height} 0 0,clip]{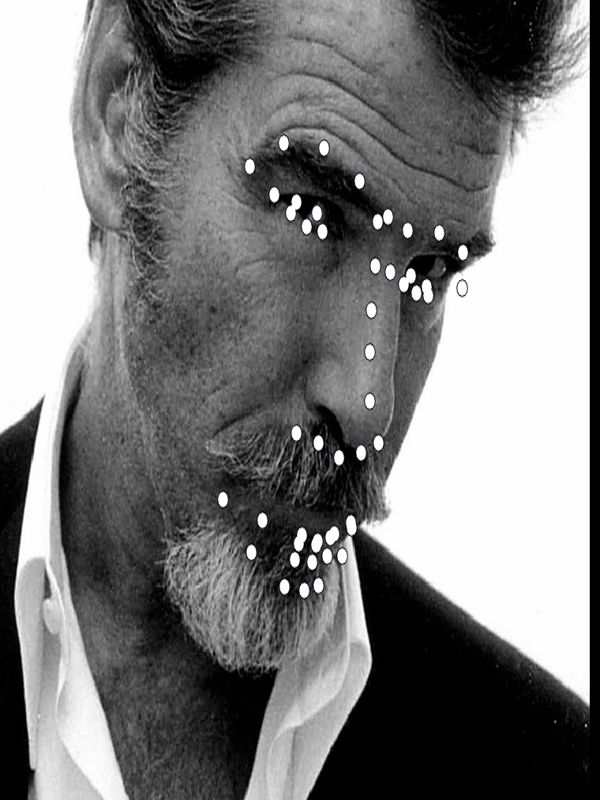}& 
\adjincludegraphics[width=0.115\textwidth,height=.1125\textwidth,trim=0 {0.1\height} 0 0,clip]{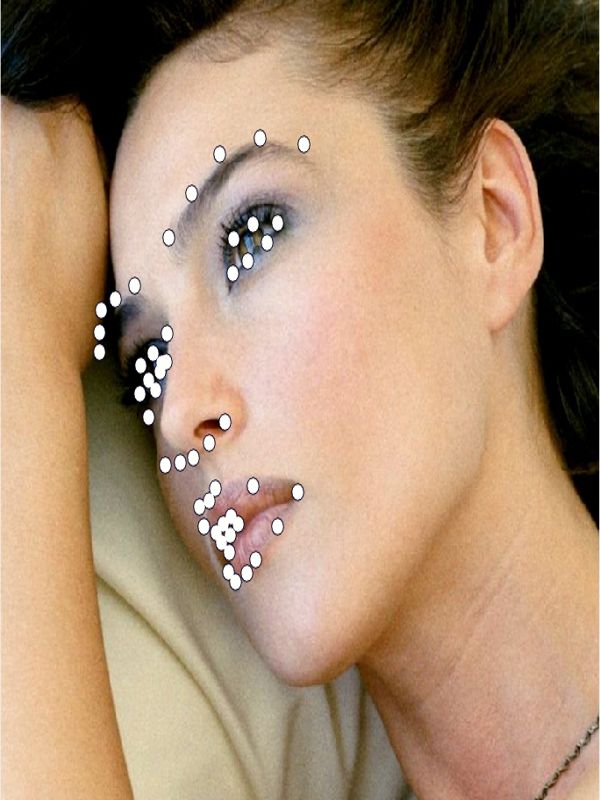}& 
\adjincludegraphics[width=0.115\textwidth,height=.1125\textwidth,trim=0 {0.1\height} 0 0,clip]{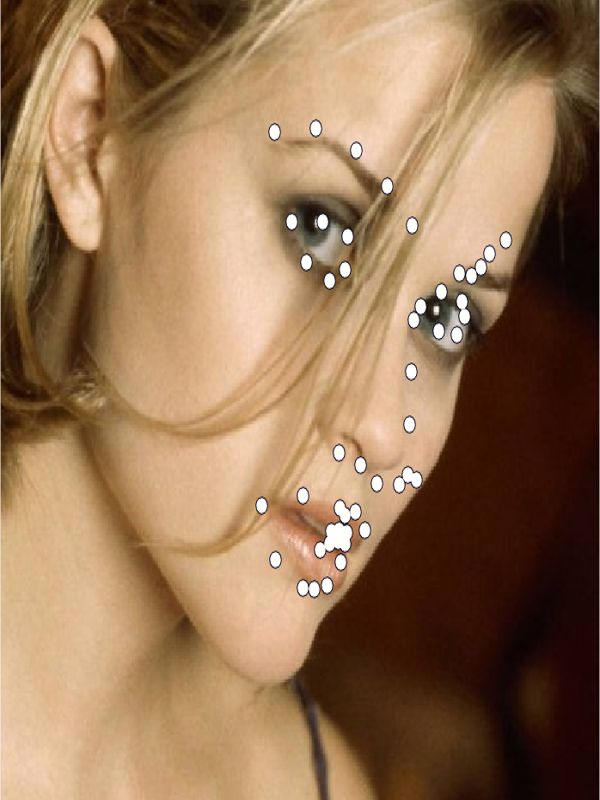}& 
\adjincludegraphics[width=0.115\textwidth,height=.1125\textwidth,trim=0 {0.1\height} 0 0,clip]{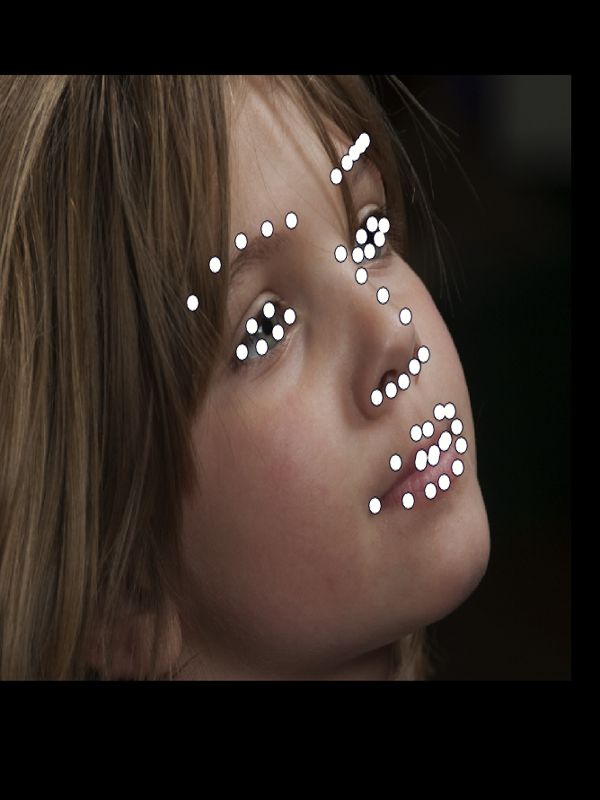}& 
\adjincludegraphics[width=0.115\textwidth,height=.1125\textwidth,trim=0 {0.1\height} 0 0,clip]{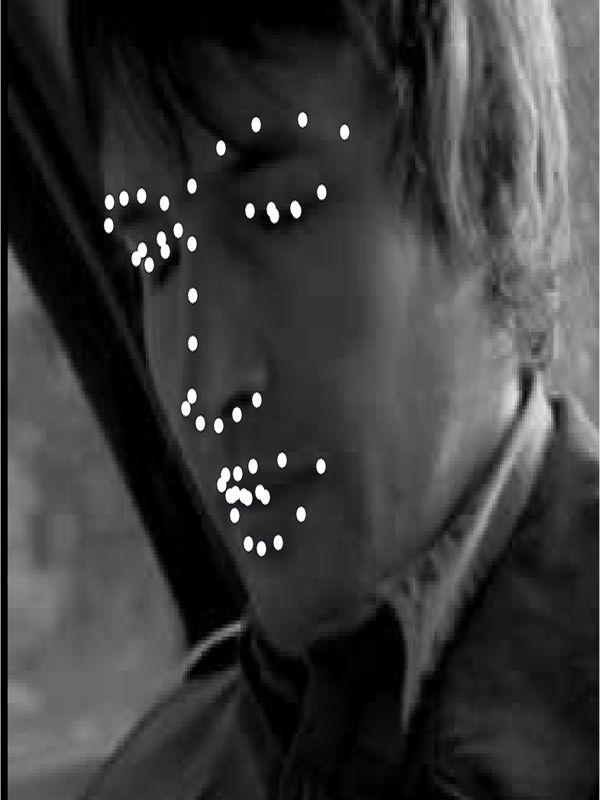}& 
\adjincludegraphics[width=0.115\textwidth,height=.1125\textwidth,trim=0 {0.1\height} 0 0,clip]{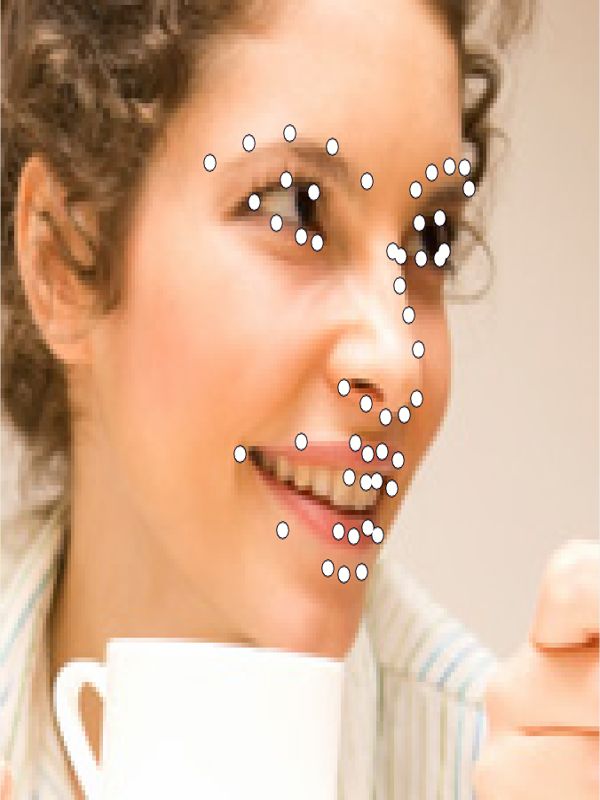}\\ 
\adjincludegraphics[width=0.115\textwidth,height=.1125\textwidth,trim=0 {0.1\height} 0 0,clip]{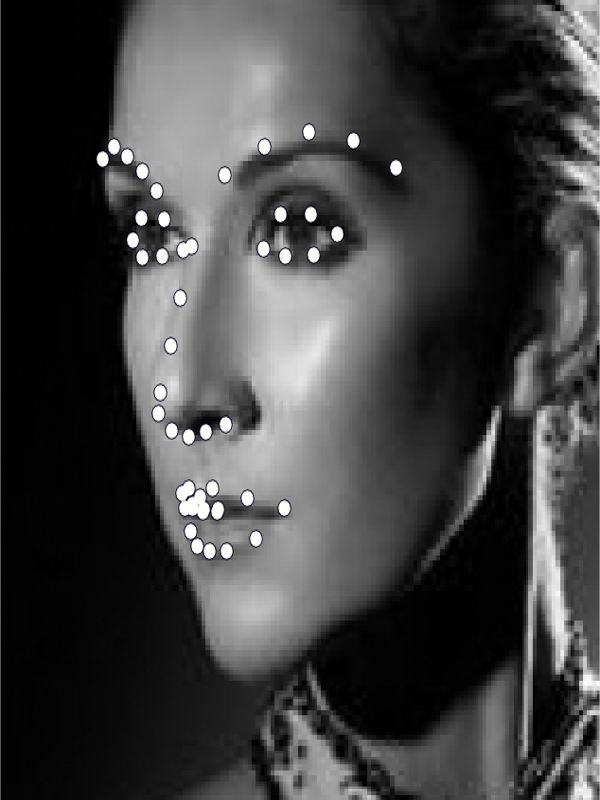}& 
\adjincludegraphics[width=0.115\textwidth,height=.1125\textwidth,trim=0 {0.1\height} 0 0,clip]{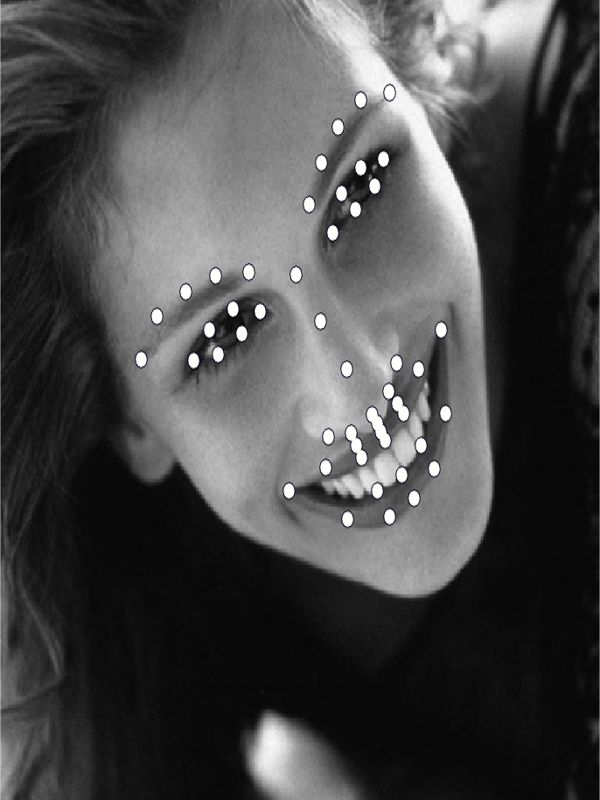}& 
\adjincludegraphics[width=0.13\textwidth,height=.1125\textwidth,trim=0 {0.1\height} 0 0,clip]{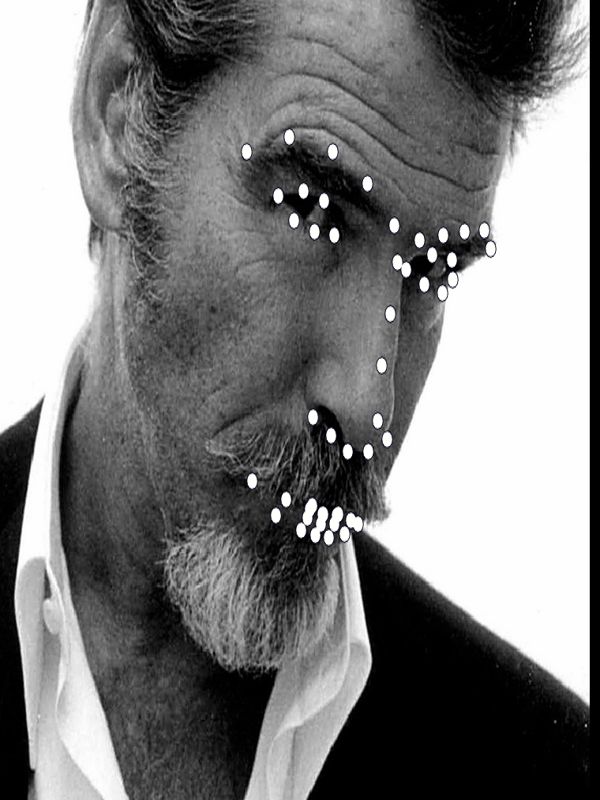}& 
\adjincludegraphics[width=0.115\textwidth,height=.1125\textwidth,trim=0 {0.1\height} 0 0,clip]{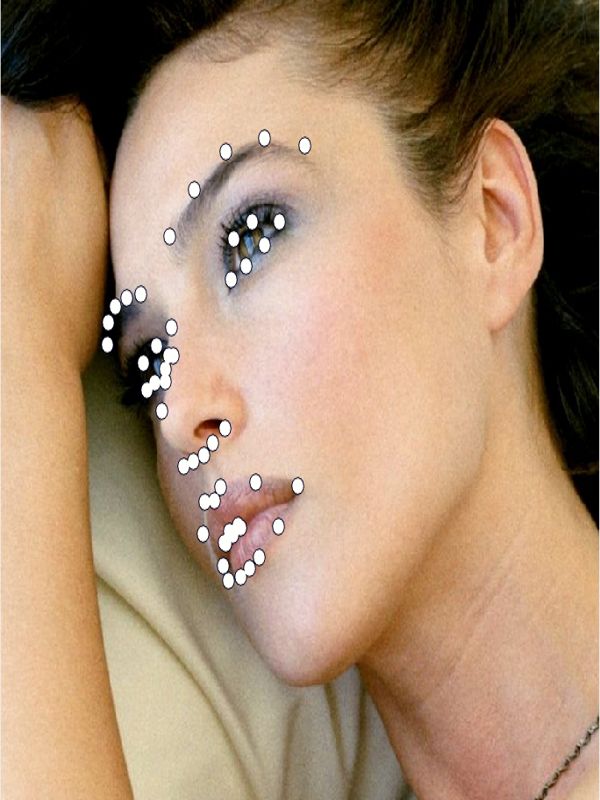}& 
\adjincludegraphics[width=0.115\textwidth,height=.1125\textwidth,trim=0 {0.1\height} 0 0,clip]{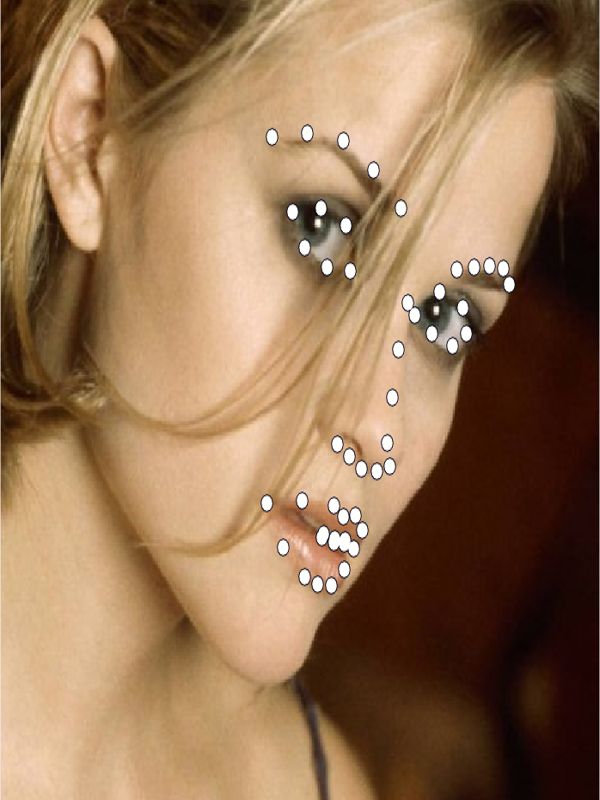}& 
\adjincludegraphics[width=0.115\textwidth,height=.1125\textwidth,trim=0 {0.1\height} 0 0,clip]{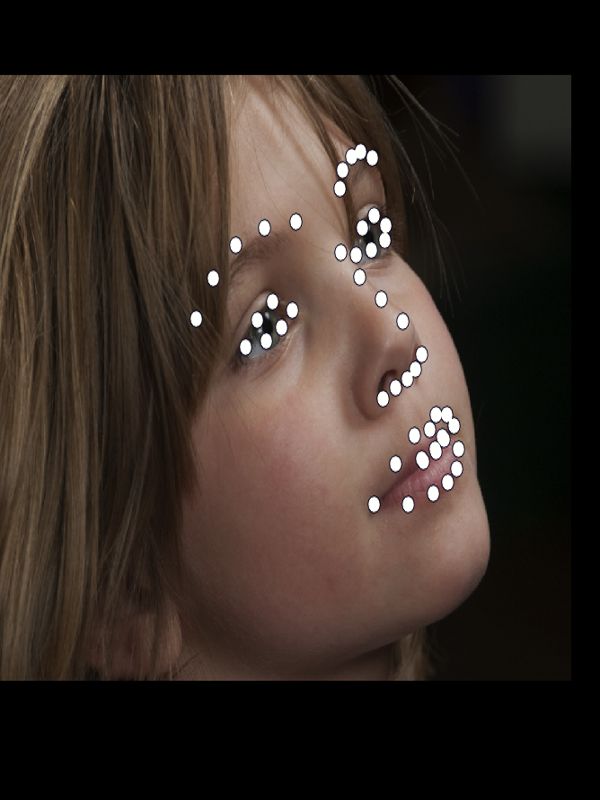}& 
\adjincludegraphics[width=0.115\textwidth,height=.1125\textwidth,trim=0 {0.1\height} 0 0,clip]{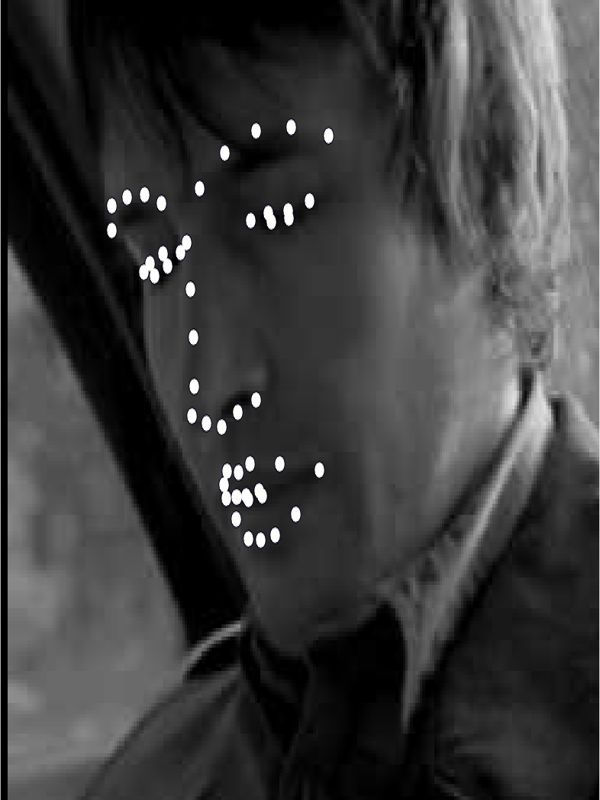}& 
\adjincludegraphics[width=0.115\textwidth,height=.1125\textwidth,trim=0 {0.1\height} 0 0,clip]{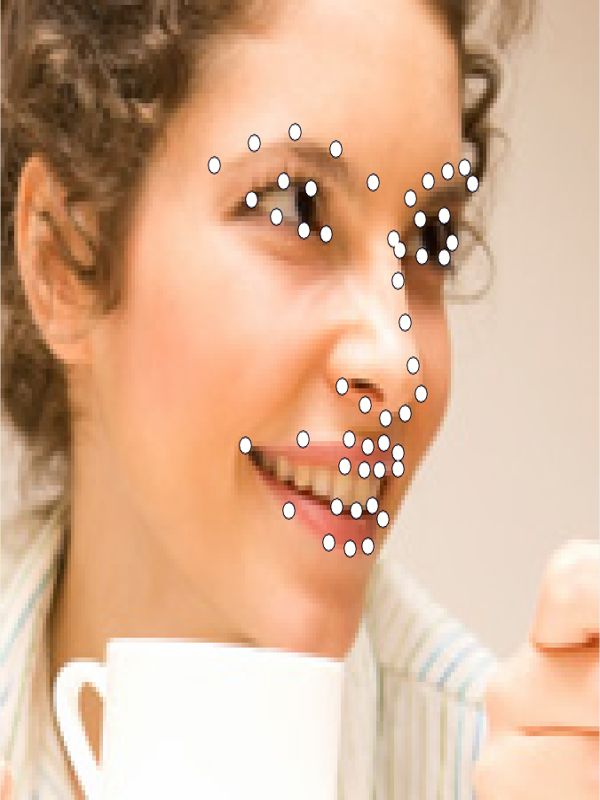}\\[-8pt]
\end{tabular}
 \caption{Visual results on the challenging subset of 300W  dataset. {\em First row:} SDM.  {\em Second row:} TI-SDM. {\em Third row:} Our MIX algorithm. Transformation invariance (TI) significantly improves the accuracy of SDM. Improvement from single models (SDM and TI-SDM) to mixture models (MIX) is apparent particularly for large out-of-plane rotations and unusual facial  expressions.}
\label{fig:ImageComparison300W} 
\end{figure*}

In the second experiment, we compare several variants of our algorithm and analyze the contribution of each of the novel components described in Section~\ref{sec:MoE}. 
The baseline algorithm for this experiment is SDM~\cite{xiong2013supervised}. MIX($L$) refers to our Mixture of Invariant Experts with $L$ experts (Section~\ref{sec:mixture}), and {\it with} or {\it without const.} refers to whether or not we use our extended deformation-constraint features (Section~\ref{sec:deformation}). \mbox{TI-SDM} is our Transformation-Invariant SDM (Section~\ref{sec:transformationInvariance}), which could also be called MIX(1) w/out const. Fig.~\ref{fig:CurvesAFWandvariations} (right) shows that each element of our algorithm improves its performance. Performance is significantly improved by adding transformation invariance (SDM $\rightarrow$ TI-SDM), by including the mixture of experts at each stage of the cascade (TI-SDM $\rightarrow$ MIX(3) w/out const.), and by using the extended deformation constraint features (MIX(3) w/out const. $\rightarrow$ MIX(3) with const.).  Using mixtures of more than 3 regression experts (MIX(3) $\rightarrow$ MIX(5)) yields very minor improvement. This is because of the limited number of training images, particularly with extreme expressions or large out-of-plane rotations, which leads experts specializing in these less common face shapes to overfit the data (as we observed during cross-validation).
In Fig.~\ref{fig:ImageComparison300W}, we visually compare sample results on the challenging subset of 300W dataset. The improvement from a cascade of single models (SDM and TI-SDM, rows 1--2) to a cascade of mixture models (MIX, row 3) is greatest for large out-of-plane rotations and unusual facial expressions. As shown in Fig.~\ref{fig:MixtureAndClusters}, each expert specializes for particular poses and expressions, yielding more precise alignment.

In the third experiment, we illustrate the behavior of deformation constraint features by simulating a case in which a few points are poorly initialized or drift away during any regression stage. As shown in column 1 of Fig.~\ref{fig:DeformationConstraint}, we initialize the alignment algorithm within the detection bounding box as usual, but to simulate drifting points we manually displace the two points shown in red (on the left eyebrow and on the outer corner of the right eye) to outside of the detection box. We then align using two models, one without deformation constraint (column 2) and the other with our extended deformation-constraint features (column 3).  The model without deformation constraint fails to correct the outlier points, whereas the deformation constraint features move outlier points towards the prototype shape of the expert, enabling it to obtain the correct landmark locations.

On a single core of an Intel Core i5-6600 3.30GHz processor, MIX with 3 experts runs at $65$ ms, of which SIFT feature computation takes $54$ ms and the rest of the algorithm takes $11$ ms. With multi-core implementation (3 experts run in parallel), run time is reduced to $30$ ms (including SIFT feature computation).


%



%
%
%
\section{Conclusion} \label{sec:discuss}

We proposed a novel face alignment algorithm based on a cascade in which each stage consists of a mixture of transformation-invariant (e.g., affine-invariant) regression experts. Each expert specializes in a different part of the joint space of pose and expressions by (affine) transforming the landmark locations to its prototype shape and learning a customized regression model. We also present a method to include deformation constraints within the discriminative alignment framework.  Extensive evaluation on benchmark datasets shows that the proposed method significantly improves upon the state of the art.

\end{document}